\pdfoutput=1

\documentclass[11pt]{article}

\usepackage{ACL2023}

\usepackage{booktabs}
\usepackage{color}
\usepackage{times}
\usepackage{latexsym}
\usepackage{amsmath}
\usepackage{float} 
\usepackage{fontawesome}
\usepackage{latexsym}
\usepackage{graphicx}
\usepackage{multirow}
\usepackage{amsmath}
\usepackage{booktabs}
\usepackage{makecell}
\usepackage{listings}
\usepackage{multirow}
\usepackage{amsfonts}
\usepackage{dsfont}
\usepackage{tikz, pgfplots}
\usepackage{bbding}
\usepackage{arydshln}
\usepackage{xspace}
\usepackage{xurl}
\usepackage{enumitem}

\definecolor{myblue}{RGB}{65,116,177}
\definecolor{mygreen}{RGB}{94,129,63}
\definecolor{myorange}{RGB}{184,97,41}

\usepackage[T1]{fontenc}

\usepackage[utf8]{inputenc}

\usepackage{microtype}

\usepackage{inconsolata}

\newcommand{\rc}{\textsc{RefChecker}\xspace}

\title{\rc: Reference-based Fine-grained Hallucination Checker and Benchmark for Large Language Models}

\author{Xiangkun Hu\textsuperscript{1}, Dongyu Ru\textsuperscript{1}, Lin Qiu\textsuperscript{1}, Qipeng Guo\textsuperscript{2}, Tianhang Zhang\textsuperscript{1}\\\textbf{Yang Xu\textsuperscript{3}, Yun Luo\textsuperscript{4}, Pengfei Liu\textsuperscript{3}, Yue Zhang\textsuperscript{4} and Zheng Zhang\textsuperscript{1}}\\
\textsuperscript{1}Amazon AWS AI\quad
\textsuperscript{2}Shanghai AI Lab\\
\textsuperscript{3}Shanghai Jiaotong University\quad
\textsuperscript{4}Westlake University\\
  {\tt \{xiangkhu, rudongyu, quln, zzthang, zhaz\}@amazon.com}, {\tt guoqipeng@pjlab.org.cn}\\
    {\tt \{xuyang0112, pengfei\}@sjtu.edu.cn}, {\tt \{luoyun, yue.zhang\}@wias.org.cn}
}

\begin{document}
\maketitle
\begin{abstract}
Large Language Models (LLMs) have shown impressive capabilities but also a concerning tendency to hallucinate. This paper presents \rc, a framework that introduces \emph{claim-triplets} to represent claims in LLM responses, aiming to detect fine-grained hallucinations. In \rc, an extractor generates claim-triplets from a response, which are then evaluated by a checker against a reference. We delineate three task settings: Zero, Noisy and Accurate Context, to reflect various real-world use cases. We curated a benchmark spanning various NLP tasks and annotated 11k claim-triplets from 2.1k responses by seven LLMs. \rc supports both proprietary and open-source models as the extractor and checker. Experiments demonstrate that claim-triplets enable superior hallucination detection, compared to other granularities such as response, sentence and sub-sentence level claims. \rc outperforms prior methods by 6.8 to 26.1 points on our benchmark and the checking results of \rc are strongly aligned with human judgments\footnote{This work is open sourced at \url{https://github.com/amazon-science/RefChecker}}.
\end{abstract}

\section{Introduction}
\label{sec:intro}

Large Language Models (LLMs) have sparked a revolution in Natural Language Processing (NLP), covering diverse tasks with a unified architecture~\cite{LLMSurvey}. However, LLMs exhibit a tendency to generate hallucinated contents that can be difficult to discern, posing a potential risk of misleading users.~\cite{huang2023survey}. Hallucination detection is therefore an important task~\cite{manakul-etal-2023-selfcheckgpt, min-etal-2023-factscore, chern2023factool}. 

\begin{figure}
    \centering
    \includegraphics[width=0.9\linewidth]{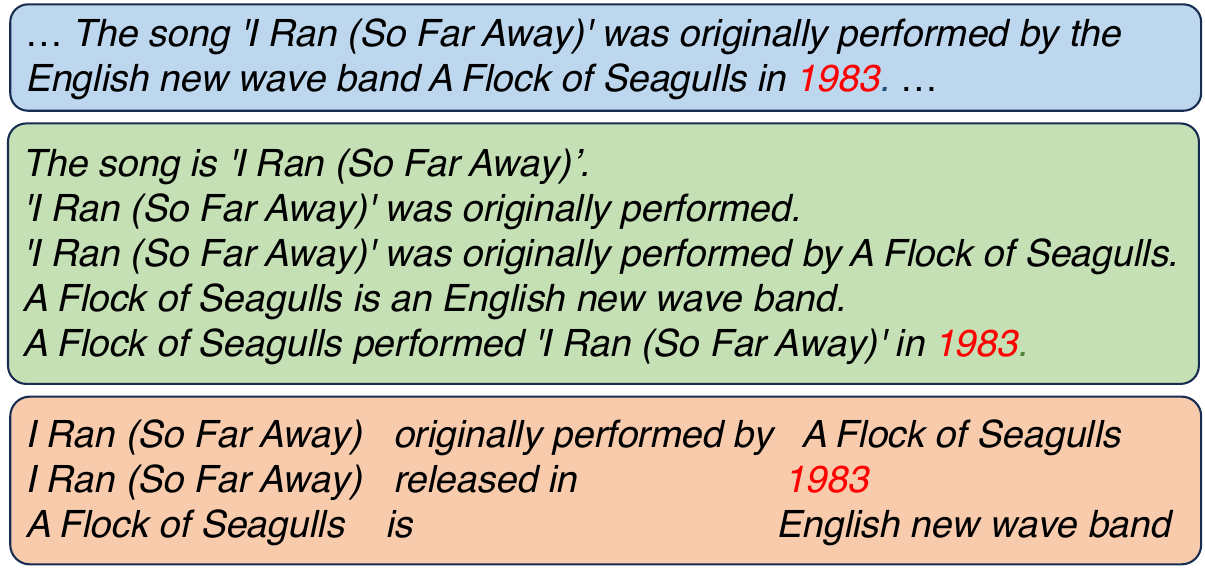}
    \caption{An example response split into \textcolor{myblue}{sentence}, \textcolor{mygreen}{sub-sentence}~\cite{min-etal-2023-factscore}, \textcolor{myorange}{triplets}, and the hallucination \textcolor{red}{\emph{1983}}. Triplets define the boundary of claims more clearly, are fine-grained and covers non-overlapping facts (unlike sub-sentences).}
    \label{fig:gran_case}
    \vspace{-15pt}
\end{figure}

\begin{figure*}[t]
    \centering
    \includegraphics[width=0.9\textwidth]{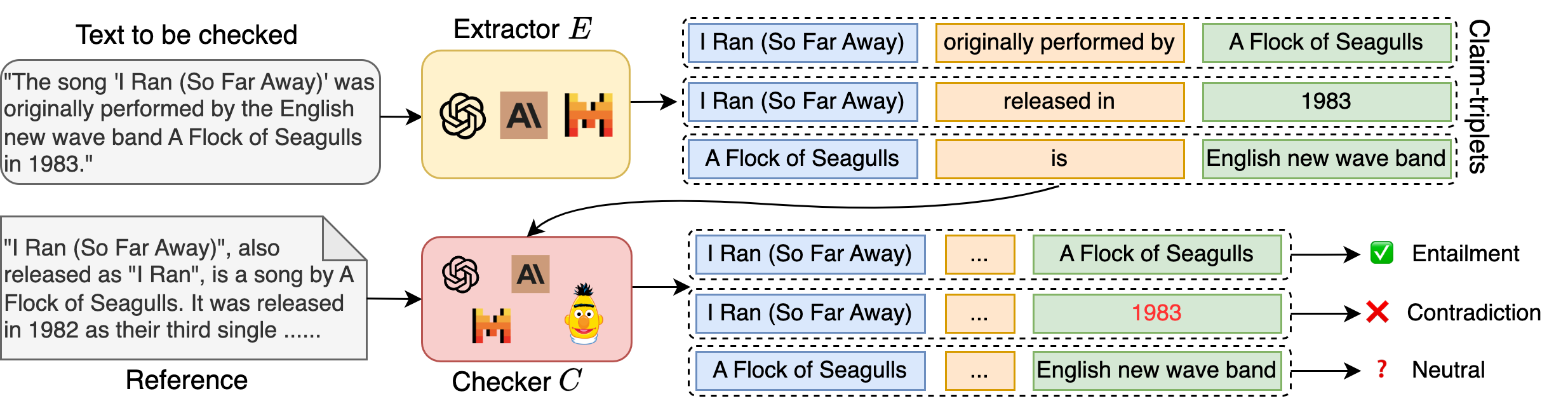}
    \caption{The \rc framework comprises two main components: an extractor denoted as $E$ and a checker denoted as $C$. Given a text to be checked, typically a response generated by an LLM, the extractor takes it as input and generates a set of knowledge triplets, referred to as claim-triplets. Subsequently, the checker assesses each claim-triplet by comparing it against a reference, assigning a hallucination label based on the evaluation.}
    \label{fig:refchecker_pipeline}
    \vspace{-15pt}
\end{figure*}

Detecting hallucination is essentially a job of comparing a response against a reference. However, several challenges remain: determining the appropriate unit of analysis for comparison, building a comprehensive benchmark reflecting real-world LLM applications, developing a unified, automated framework that scales detection across diverse tasks. 
In this work, we introduce \rc to tackle these challenges.

In terms of checking granularity, response level checking~\cite{lin-etal-2022-truthfulqa, li-etal-2023-halueval} suffices if the query and response is about a simple fact. However, when responses are complex and long, response-level checking is not only uninformative but can also cause false-negative when hallucination is local. This is common in real-world use cases, for example, the response from Llama 2~\cite{touvron2023llama} in our benchmark (described later) contains 150 tokens on average. For fine-grained detection, \citet{manakul-etal-2023-selfcheckgpt} takes the sentences in a response as the checking units. \citet{min-etal-2023-factscore} and \citet{chern2023factool} further extracts short phrases (we term them as sub-sentences) as the claims, as one sentence may contain multiple hallucinations, or one hallucination may span across sentence boundaries. However, sub-sentences are structurally difficult to define, making it challenging to form high-quality demonstrations to be used by LLMs with in-context learning. To this end, we propose to extract knowledge triplets as checking units. We show an example with different granularity in Figure~\ref{fig:gran_case}. Triplets exhibit fine-grained and clearly separated semantics. These triplets are called \emph{claim-triplets}. Experiments show that checking with claim-triplets gains 4 to 9 points of improvement over other granularity on our benchmark (cf. Sec.~\ref{sec:compare_other_granularities}).

For better evaluation, we curate a comprehensive dataset on which we can benchmark hallucination under different context quality and availability.
Using this benchmark, we conducted human evaluation on 2,100 responses from 7 LLMs. We annotated 11k claim-triplets with 95\% Inter-Annotator Agreement on 23\% of the annotations. Compared with recent proposed benchmarks~\cite{manakul-etal-2023-selfcheckgpt, min-etal-2023-factscore, chern2023factool}, it covers a more diverse range of domains and tasks, with more LLMs and responses evaluated (see Table~\ref{tab:compare_related_work}). As expected, we found by human evaluation that hallucination is the most pronounced (cf. Appendix~\ref{appendix:observation_from_human_eval}) when LLMs are asked to generate responses solely from its memory (Zero Context), followed by responding to noisy references in RAG (retrieval augmented generation) setting~\cite{shuster-etal-2021-retrieval-augmentation} (Noisy Context)  and finally when references are more or less noisy free (Accurate Context). 

\rc (Figure~\ref{fig:refchecker_pipeline}) is a fully automated framework that scales hallucination detection across different tasks. The extractor generates claim-triplets from the response and the checker evaluates each of the claim-triplets by comparing them with the reference. In contrast to recent work that only differentiates factual and non-factual claims, the checker in \rc also considers unverifiable claims when the reference is insufficient for checking. Both the extractor and checker supports proprietary (e.g. GPT-4~\cite{openai2023gpt4} or Claude 2~\cite{anthropic_claude}) and open-source models (e.g. Mistral~\cite{jiang2023mistral} and RoBERTa~\cite{liu2019roberta} based models). We made careful study to select and recommend configurations that give results consistent with human annotation, and show 6.8 to 26.1 points of improvement over the best alternative (Sec.~\ref{sec:compare_other_approach}).

Our key contributions include:

\begin{itemize}
    \item \textbf{Claim-triplet formulation:}  Our novel ``claim-triplet'' analysis outperforms existing methods by up to 9 points, pinpointing factual inconsistencies within responses. 
    \item \textbf{Comprehensive benchmark:} We developed a robust benchmark covering three classes of real-world LLM tasks with 11,000 manually annotated claim-triplets across 7 LLMs.
    \item \textbf{Automatic checking framework:} Our \rc framework extracts and verifies claim-triplets, boosting consistency by 19-26 points over prior methods and works with both proprietary and open-source models. 
\end{itemize}

\section{Related Work}
\label{sec:related_work}

\begin{table*}[ht]
    \centering
    \scalebox{0.6}{
\begin{tabular}{lcccccccc}
\toprule
\multirow{2}{*}{\textbf{Method}} & \multirow{2}{*}{\textbf{Context Setting}} & \multicolumn{2}{c}{\textbf{Claim Extraction}} & \multicolumn{2}{c}{\textbf{Checking}} & \multicolumn{3}{c}{\textbf{Benchmark}} \\ 
\cline{3-9}
  & & \textbf{Claim} & \textbf{Extractor} & \textbf{Label} & \textbf{Checker} & \textbf{Domain} & \textbf{Task} & \textbf{Evaluated Responses}\\ 
\hline
SelfCheckGPT & Zero Context & Sentence & - & 3-way  & GPT & Wikipedia & Bio Generation & 238 from GPT-3 \\ \hdashline
FActScore & Zero Context & Sub-sentence & GPT & Binary & GPT & Wikipedia & Bio Generation & 505 from 3 LLMs \\ \hdashline
FacTool & Zero Context & \makecell{Sub-sentence\\Snippet\\Statement\\Tuple} & GPT & Binary & GPT & \makecell{Wikipedia\\Python\\Math\\Sci. Text} & \makecell{QA\\Code Generation\\Math Problems\\Sci. Review} & 514 from ChatGPT \\ 
\hline
\rc & \makecell{Zero Context\\Noisy Context\\Accurate Context} & Triplet & \makecell{GPT\\ Claude\\ Mixtral*\\ Mistral*} & 3-way & \makecell{GPT\\ Claude\\ AlignScore* \\ NLI*\\ RepC*} & \makecell{Wikipedia\\ Web} & \makecell{QA\\ RAG\\ Summarization\\ IE} & 2,100 from 7 LLMs \\
\bottomrule
\end{tabular}
}
    \caption{A comparison of \rc with previous approaches for hallucination detection. The ``*'' symbols alongside the extractors and checkers indicate that these models are open-sourced. \rc uses triplets as the claims instead of sentences or sub-sentences. The \rc benchmark covers more context settings and more diverse tasks. The human evaluation covers more LLMs and responses. \rc pipeline supports both proprietary and open-source models, facilitating broader adoption across various applications.}
    \label{tab:compare_related_work}
    \vspace{-15pt}
\end{table*}

We undertake a review of prior work relevant to our study and compare them with \rc. The comparative analysis with three representative methods is encapsulated in Table~\ref{tab:compare_related_work}.

\paragraph{Hallucinations in LLMs}

Hallucinations frequently occur in NLP tasks like summarization~\cite{maynez-etal-2020-faithfulness, cao-etal-2022-hallucinated}, machine translation~\cite{10.1162/tacl_a_00615, guerreiro-etal-2023-looking}, dialog systems~\cite{honovich-etal-2021-q2, dziri2022faithdial} and RAG~\cite{shuster-etal-2021-retrieval-augmentation}. According to a recent comprehensive survey~\cite{huang2023survey}, hallucinations can be categorized to factuality hallucinations and faithfulness hallucinations. Factuality hallucinations involve claims contradicted by real-world facts, while faithfulness hallucinations are inconsistent with the input content. Recent research on hallucination detection primarily concentrates on factuality hallucinations, such as SelfCheckGPT~\cite{manakul-etal-2023-selfcheckgpt}, FActScore~\cite{min-etal-2023-factscore} and FacTool~\cite{chern2023factool}. We address both factuality and faithfulness hallucinations and further categorizing them into three contextual settings to align with real-world use cases.

\paragraph{Granularity of Claims}

Claims are pivotal for evaluating responses generated by LLMs. Response level checking~\cite{lin-etal-2022-truthfulqa, li-etal-2023-halueval} is too coarse-grained for long-form responses. For fine-grained detection, sentence level~\cite{manakul-etal-2023-selfcheckgpt} and sub-sentence level checking~\cite{min-etal-2023-factscore, chern2023factool} have been proposed. However, these approaches still face limitations, as discussed in Sec.~\ref{sec:intro}. In this paper, we employ knowledge triplets extracted from responses as claims which provide a structured framework for defining claim granularity.

\paragraph{Hallucination Checking}

One line of work for hallucination checking focus on resource-free checking with no requirements of reference. They mainly depend on self-contradiction~\cite{mündler2023selfcontradictory} or uncertainty~\cite{zhang-etal-2023-enhancing-uncertainty} of the LLMs, or self-consistency between randomly sampled responses~\cite{manakul-etal-2023-selfcheckgpt, chen2023quantifying, zhang-etal-2023-sac3}. The effectiveness of these methods depends on the LLM-based checker's capability including knowledge coverage, instruction following ability and calibration. Furthermore, the self-consistency based methods needs to sample several responses for cross checking, which is costly. \rc is aligned with another line of work which requires references to check with~\cite{min-etal-2023-factscore, chern2023factool}. 
In addition, \rc adopts a 3-way classification framework to cover unverifiable claims as opposed to the binary classification used in previous work, which can only distinguish factual and non-factual claims. Moreover, we have introduced open-source solutions for both claim extraction and verification, diverging from the predominant reliance on proprietary LLMs in prior research. 


\paragraph{Hallucination Detection Benchmarks}

The existing benchmarks for hallucination detection primarily focus on response-level detection~\cite{lin-etal-2022-truthfulqa, yang2023new}, or limited to specific domains and tasks~\cite{manakul-etal-2023-selfcheckgpt, min-etal-2023-factscore}, or solely address factuality hallucinations~\cite{chen2023felm, chern2023factool, Wang2023FactcheckGPTEF}. In contrast, our proposed benchmark offers a broader scope, encompassing a diverse range of tasks and domains. Moreover, our human evaluation process involves a more extensive examination of various LLMs with more responses.

\section{\rc: Definition and Benchmark}
\label{sec:taxonomy}

\begin{figure*}[ht]
    \centering
    \includegraphics[width=0.9\textwidth]{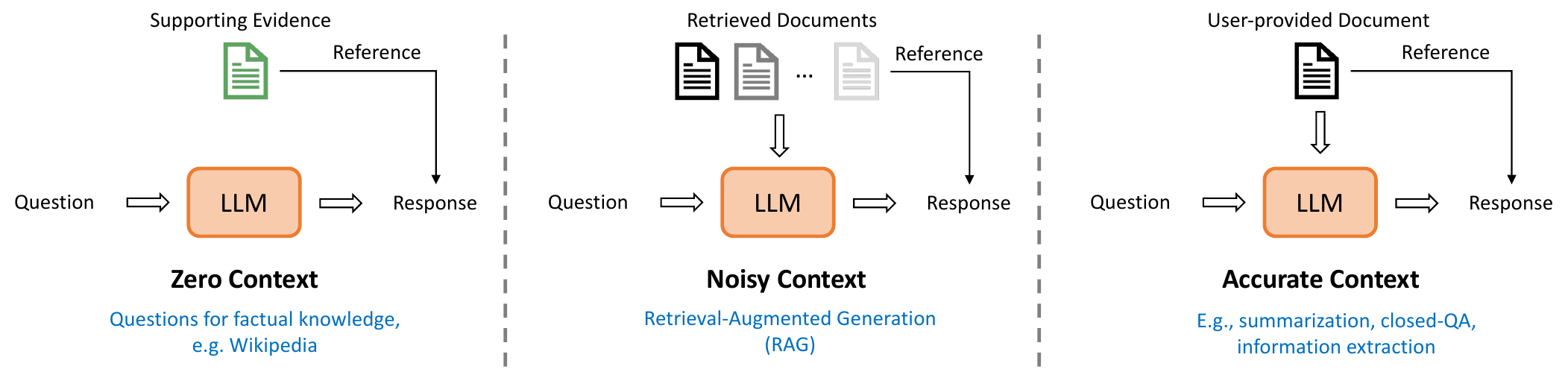}
    \caption{Illustration of three settings of context, tasks and references. Zero Context is about seeking factual knowledge from the internal memory of the LLMs. Noisy Context has context information retrieved from a knowledge source, which is a RAG use case. Accurate Context has context provided in the input prompt. For Noisy and Accurate Context, we take the input context as the reference.}
    \label{fig:task_settings}
    \vspace{-15pt}
\end{figure*}

Hallucinations are claims made by LLMs not supported by factual knowledge, which we refer to as references; detecting hallucinations involves comparing the claims against the references. This process depends on context settings, granularity of checking and how we categorize the hallucinations. We will discuss them in turn.

\subsection{Context Settings and Benchmarks}

We differentiate three context settings covering various tasks and employ different benchmarks for each setting.

\paragraph{Zero Context (ZC)} Tasks in this setting require the LLM to respond solely based on its internal knowledge. Therefore, in principle, references should be in the training corpus. In practise, we use NaturalQuestions (NQ)~\cite{kwiatkowski-etal-2019-natural} as proxy dataset to represent such stable knowledge. The knowledge seeking questions from NQ serve as prompts to perform the closed-book question answering task, and the annotated long answer as the reference for a fair evaluation of LLMs and comparison between different approaches. These references are paragraphs from Wikipedia articles, which are widely used for pre-training LLMs~\cite{LLMSurvey}, making them a suitable proxy for the internal knowledge of these LLMs.  

\paragraph{Noisy Context (NC)} In this setup, the LLM receives additional context retrieved from some external knowledge source, which may contain noisy or irrelevant information. NC is also known as RAG, a crucial use case frequently encountered in real-world applications. We utilize questions sourced from MS MARCO~\cite{DBLP:conf/nips/NguyenRSGTMD16}. Each question in this dataset is accompanied by a list of documents retrieved from the internet, serving as the input context. 

\paragraph{Accurate Context (AC)} This setting is similar to NC but the reference is typically noise-free. Examples include text summarization, closed-QA and information extraction tasks. We employ a subset from the databricks-dolly-15k~\cite{DatabricksBlog2023DollyV2} instruction tuning dataset which covers the 3 tasks mentioned before.

\begin{figure}
    \centering
    \includegraphics[width=0.9\linewidth]{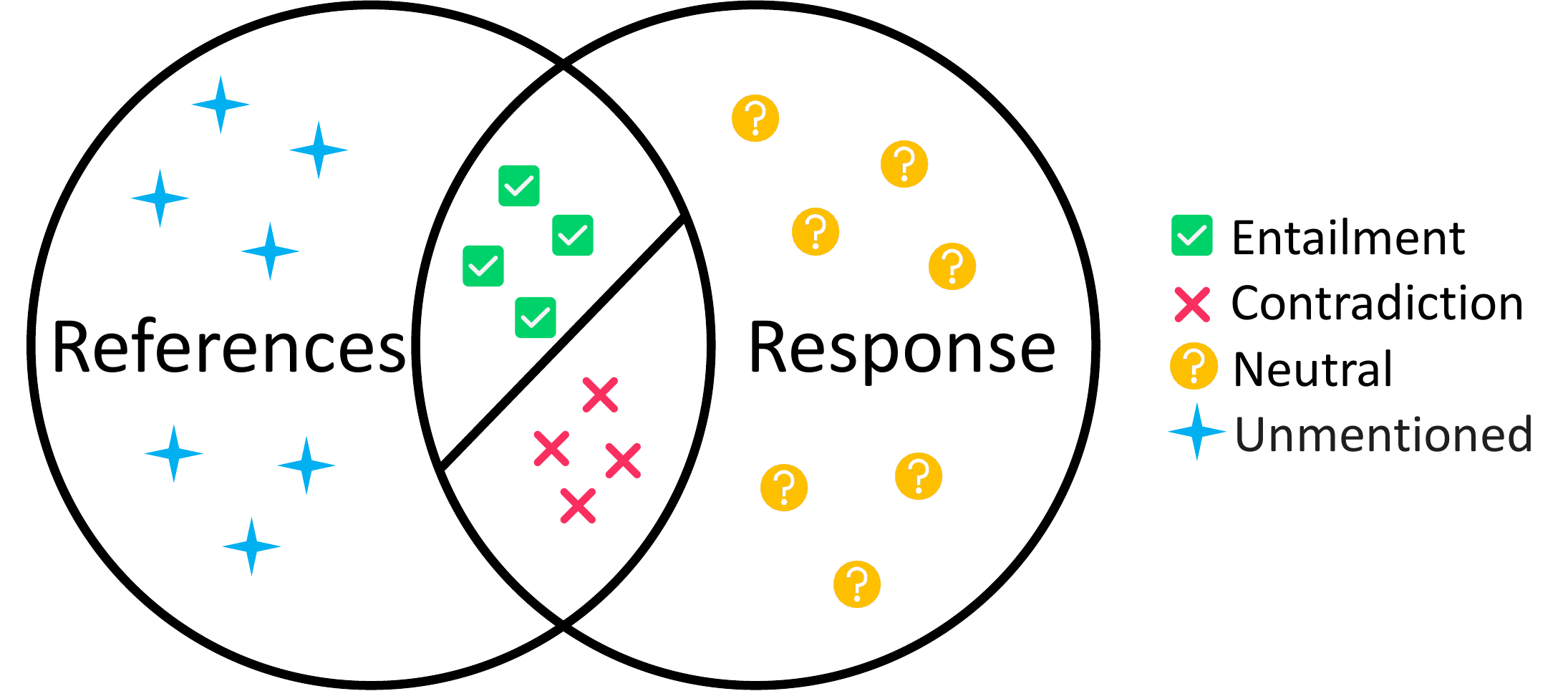}
    \caption{Definition of fine-grained hallucinations in an LLM-generated response compared with references. The intersections of the response and the references are the claims either supported (Entailment) or refuted (Contradiction) by the references. The remaining parts in the response are claims not verifiable by the references (Neutral). The other parts of the references are the content not mentioned in the response.}
    \label{fig:halu_definitions}
    \vspace{-15pt}
\end{figure}

For both Noisy and Accurate Context settings, the prompt is the reference followed by the query, whereas for Zero Context, the prompt is just the query (Figure~\ref{fig:task_settings}). The benchmark contains 300 examples in total, 100 for each setting, and we summarize it in Table~\ref{tab:benchmark_data}. The details of the benchmark curation process are described in Appendix~\ref{appendix:data}.


\subsection{Claim-Triplets and Definition of Hallucination}
\label{sec:claim_triplet}

Informally, claims are the units for the checking. This work explores the approach of representing claims with knowledge triplets. This concept is inspired by the field of knowledge graph studies, where triplets are employed to encapsulate factual knowledge units. Knowledge triplets adopt a \texttt{(head\_entity, relation, tail\_entity)} structure to capture fine-grained information within the response. We call the triplet-format claims as \emph{claim-triplets}, examples of which are shown in Figure~\ref{fig:refchecker_pipeline}.

Subsequently, the claim-triplets are compared with a reference to determine the type of hallucinations, as illustrated in Figure~\ref{fig:halu_definitions}. If a claim-triplet can be directly inferred from the reference, we classify it as \textbf{Entailment}. Conversely, if it contradicts the information in the reference, it is labeled as \textbf{Contradiction}. However, in cases where the reference is insufficient to verify the claim-triplets, we classify it as \textbf{Neutral}. In this study, we focus on verifying hallucinations in the response and do not consider unmentioned aspects in the reference, which may also be important for certain tasks.

\subsection{Human Evaluation}

We performed a human evaluation of responses generated by seven LLMs on this benchmark dataset, including GPT-4, GPT-3.5-Turbo~\cite{chatgpt}, InstructGPT (text-davinci-001)~\cite{ouyang2022training}, Claude 2, Llama 2 70B Chat, Falcon 40B Instruct~\cite{falcon40b} and Alpaca 7B~\cite{alpaca}. The process involves three steps: gathering responses, extracting claim-triplets with an extractor, and asking human annotators to evaluate these claim-triplets. We annotated a total of 11k claim-triplets for 2.1k responses. 23\% of the claim-triplets were double annotated, with 95.0\% Inter-Annotator Agreement. See Appendix~\ref{appendix:annotation} for the details of the annotation process. 

A significant finding from the human evaluation is the crucial role of contextual information for factual responses. On average, the rate of Contradiction decreased from 25\% in the absence of contextual cues (ZC) to 13\% with NC, and further reduced to 6\% with AC, which reflects the necessity of distinguishing the three context settings for separate study (cf. Appendix~\ref{appendix:observation_from_human_eval}).


\section{\rc Framework}

As illustrated in Figure~\ref{fig:refchecker_pipeline}, the \rc framework is designed as a 2-stage pipeline: an Extractor $E$ decomposes the LLM response into a set of triplets, with each of them verified by the Checker $C$. The categorization of the triplets can be optionally aggregated according to specified rules. We explain them in the subsequent subsections.

\subsection{Extractor}
\label{sec:pipeline_extractor}

Our checking framework hinges on a key assumption: the decomposition of the original text into triplets facilitates finer-grained detection and more accurate evaluation. The extraction of these triplets plays a pivotal role in achieving this objective. We apply LLMs to extract knowledge triplets from the given text. We began with GPT-4 and Claude 2 and, for both cost and efficiency concern, Mixtral 8x7B and Mistral. More specifically, we performed knowledge distillation to train a 7B Mistral-based extractor with Mixtral 8x7B as the teacher. We conducted supervised fine-tuning on 10k responses. Evaluation in Sec.~\ref{sec:exp:extractor-eval} shows competitive extraction quality of the open-source extractor. Please refer to Appendix~\ref{appendix:refchecker_extractor} for prompts used for extraction and details on extractor training.

\subsection{Checker}
\label{sec:pipeline_checker}
We experimented with two families of checkers, the first is off-the-shelf LLMs including GPT-4 and Claude 2 (see Appendix~\ref{appendix:refchecker_checker} for prompts), and the second is smaller NLI models including AlignScore \cite{zha2023alignscore} and RoBERTa-NLI.\footnote{\url{https://huggingface.co/ynie/roberta-large-snli_mnli_fever_anli_R1_R2_R3-nli}} Long references in AC/NC setting are split to fit into small context windows of these small models (e.g. 200 tokens), and the results are aggregated later.

Mistral 7B \cite{jiang2023mistral}, unlike GPT-4/Claude 2, performs poorly as a zero-shot checker, its performance improves with demonstrations but is still not satisfactory. 
However, the fact that the model weight is open gives us the opportunity to improve it by fine-tuning with NLI data. There are many options we have experimented: 1) fine-tune by adding small amount of new parameters using LoRA (LoRA-sft)~\cite{hu2021lora}, 2) attach a shallow classifier, eg. SVM, 2-layer MLP, KNN after NCA projection \cite{goldberger2004neighbourhood}, on top of the internal states of the model. We call such checker RepC (for Representation-based Classifier). Such states can be selected from one layer (layer selection, LS) or an ensemble of all layers (layer ensemble, LE). As we will report in Sec.~\ref{sec:exp:checker-eval}, RepC checkers are competitive in general.

\subsection{Aggregation}

Triplet results can be aggregated to obtain the ratio of each category, therefore gives an overall measure of hallucination distribution in a response. 
To derive the performance of a particular LLM, 
we take a macro average on Entailment/Neutral/Contradiction ratios of all responses. If a scalar is preferred, we can assign certain numeric values to the catogories, for instance $-1, 0, 1$ for contradictory, neutral and entail, respectively.

The aggregation can be customized 
and this is one of the benefits of the fine-grained hallucination design in \rc.
For instance, to compare against other response-level approaches (cf. Sec.~\ref{sec:compare_other_granularities}), we adopt a rule where the response is flagged as contradictory if any one of the claim triplet is contradictory.





\begin{table*}[!ht]\small
    \centering
    \begin{tabular}{lcccccc}
    \toprule
    & \multicolumn{2}{c}{Zero Context} & \multicolumn{2}{c}{Noisy Context} & \multicolumn{2}{c}{Accurate Context} \\
 & Pearson & Spearman & Pearson & Spearman & Pearson & Spearman \\ \hline
SelfCheckGPT & 35.40 & 43.15 & 36.31 & 32.15  & 40.23 & 32.55 \\
FActScore & 42.58 & 45.60 & 33.36 & 29.91 & 27.80 & 27.05 \\
FacTool & 59.78 & 62.57 & 46.35 & 38.69 & 31.41 & 32.82 \\ \hline
\rc \\
\textit{Claud 2 + GPT-4} & \textbf{\textcolor{cyan}{83.69}} & \textbf{\textcolor{cyan}{82.99}} & \textbf{\textcolor{cyan}{53.14}} & \textbf{\textcolor{cyan}{47.89}} & \textbf{\textcolor{cyan}{60.99}} & \textbf{\textcolor{cyan}{58.96}} \\
\textit{Mistral-SFT + AlignScore} & \textbf{\textcolor{orange}{75.81}} & \textbf{\textcolor{orange}{74.16}} & \textbf{\textcolor{orange}{53.88}} & \textbf{\textcolor{orange}{45.09}} & \textbf{\textcolor{orange}{46.34}} & \textbf{\textcolor{orange}{43.22}} \\
\bottomrule
    \end{tabular}
    \caption{Automated checking results of \rc and previous approaches. The results of \rc are from the best performing combinations (\textit{extractor + checker}) of purely proprietary (blue) and purely open-source models (orange).}
    \label{tab:compare_related_work_correlation_our_benchmark}
    \vspace{-15pt}
\end{table*}

\section{Experiments}

The major difference between \rc and other related work lies in the claim granularity. So we conduct experiments to differentiate granularity first, then evaluate the whole framework.

\subsection{Comparing with Other Granularity}
\label{sec:compare_other_granularities}

\begin{figure}[t]
    \centering
    \includegraphics[width=0.85\linewidth]{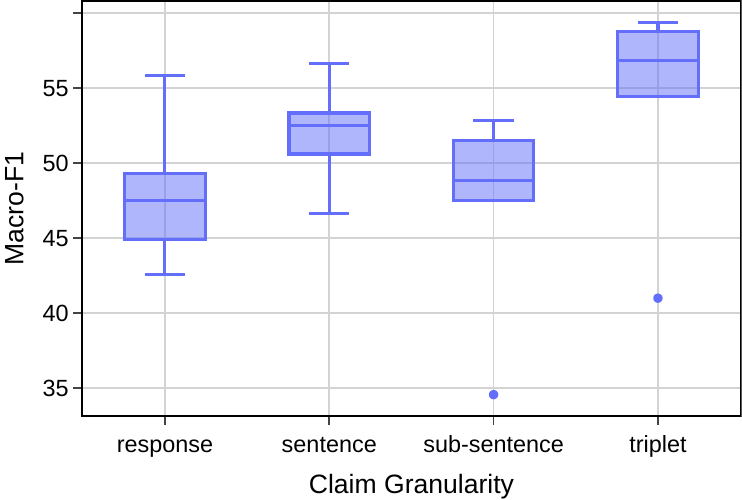}
    \caption{Performance statistics of 7 checkers under different claim granularities on 2.1k manual annotated responses. The detailed checker performance can be found in Table~\ref{tab:granularity} of Appendix~\ref{appendix:refchecker_checker}.}
    \label{fig:checker_gran}
    \vspace{-15pt}
\end{figure}

Previous works use different granularity for hallucination detection, including response, sentence and sub-sentence levels (cf. Sec.~\ref{sec:related_work} and~\ref{sec:claim_triplet}). 
We compare with them to verify the effectiveness of checking on facts with the triplet format.

To make the results with different granularities comparable to each other, we first breakdown the 2.1k annotated responses into different granularities,
then collect corresponding checker predictions respectively, and finally aggregate finer-level results all into the response-level. We utilize a strict aggregation rule with zero-tolerance on hallucinations, which means we apply max-pooling (Entailment $<$ Neutral $<$ Contradiction) over claim predictions within a response. We compare the results of 7 checkers, including 4 baseline checkers (RoBERTa-NLI,
AlignScore, GPT-4 and Claude 2) and 3 RepC-LE checkers with KNN, SVM and 2-layer MLP classifiers respectively. The evaluation metric is macro-f1 on three categories. 

As shown in Figure~\ref{fig:checker_gran}, checking at triplet-level claims is superior over other granularities, with a significant lead against response-level (10 pts macro-f1 score on average). Checking at sentence-level improves over response-level by 5 pts. However, we see a 3.5 pts drop moving to sub-sentence, one of the reasons being sub-sentence claims can overlap. Apparently, the flexibility of sub-sentences leads to poor quality of claim extraction, which subsequently affects checking.

\subsection{Comparing with Other Approaches}
\label{sec:compare_other_approach}

To contrast with other hallucination detection methodologies, we compare \rc with three popular approaches on our benchmark, SelfCheckGPT~\cite{manakul-etal-2023-selfcheckgpt}, FActScore~\cite{min-etal-2023-factscore} and FacTool~\cite{chern2023factool}. We convert metrics in these approaches to 
``hallucination rates'' as follows:
\begin{itemize}[noitemsep, leftmargin=1pt, itemindent=1pc, topsep=1pt]
    \item SelfCheckGPT: the average score of the sentences within a response. The score is 0, 0.5 and 1 for an \texttt{accurate}, \texttt{minor\_inaccurate}, and \texttt{major\_inaccurate}, respectively.
    \item FactScore/FacTool: the proportion of claims not supported (FactScore) or non-factual (FacTool) by the reference.
    \item \rc: the proportion of Neutral and Contradiction claims. 
\end{itemize}

Table~\ref{tab:compare_related_work_correlation_our_benchmark} presents the Pearson and Spearman correlations between the hallucinations rates as evaluated by human and the model results. It reveals that \rc significantly outperforms previous methods across all three context settings with both proprietary and open-source models. Specifically, the combination of Claude 2 + GPT-4 outperforms the best alternative, FacTool, by 6.8 to 26.1 points. We additionally apply \rc to SelfCheckGPT's dataset, and observe that in 11 out of the 15 combinations (73\%) of \rc outperform SelfCheckGPT (cf. Table~\ref{tab:selfcheckgpt_data_results} in Appendix~\ref{appendix:compare_other_methods}).

\begin{figure}[t]
    \centering
    \includegraphics[width=0.85\columnwidth]{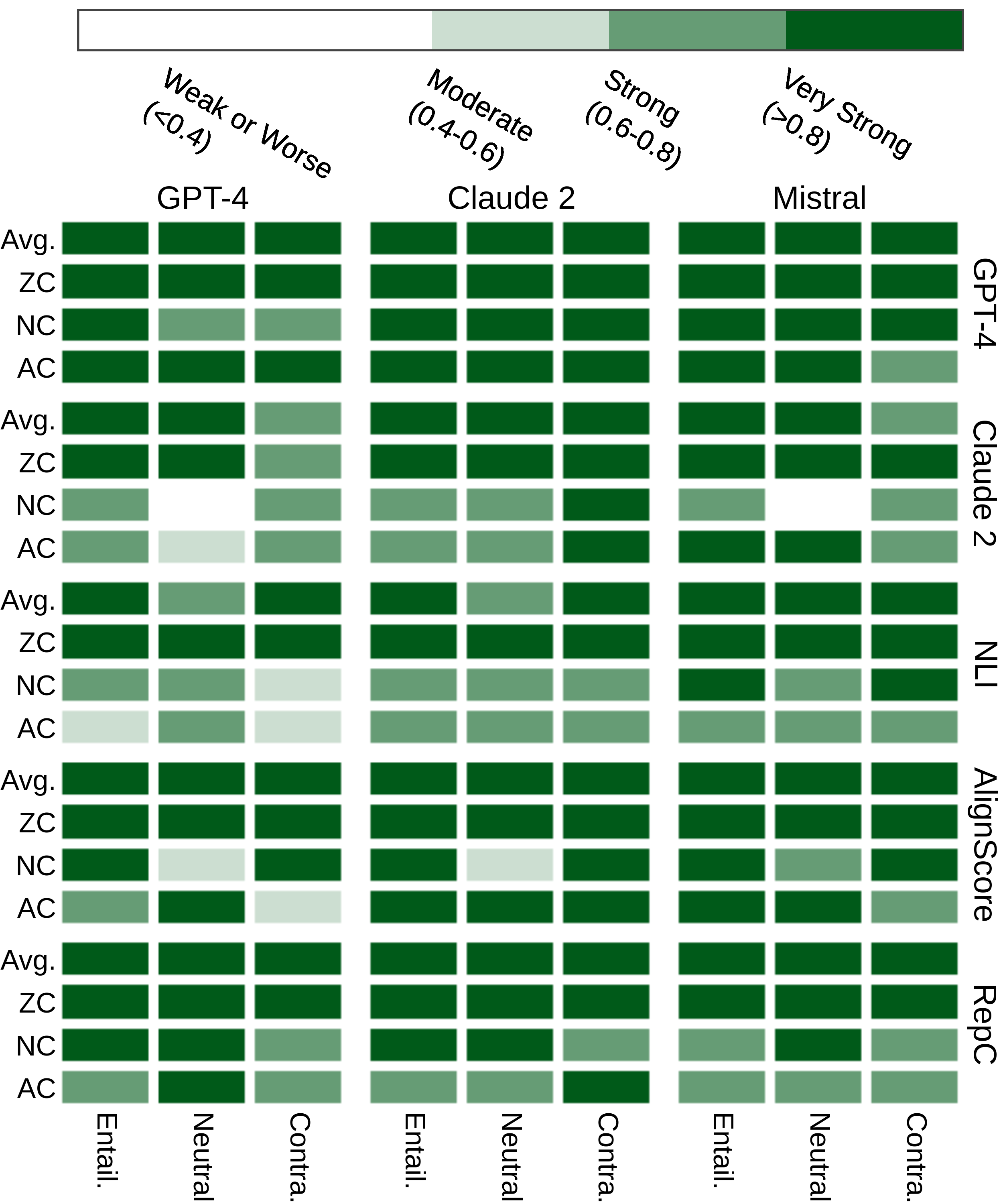}
    \caption{Spearman's rank correlation coefficients between \rc and human evaluation. Results are grouped regarding extractors and checkers used. Results for each extractor (column) and checker (row) are arranged into a sub-matrix in the figure, with values for 4 context settings (one additional for average) and 3 ranking metrics.}
    \label{fig:correlation}
    \vspace{-15pt}
\end{figure}

\subsection{Evaluation on \rc Framework}
\label{sec:pipeline-perf}

To validate the robustness and reliability of \rc, we checked 7 LLMs' responses on our benchmark and compared rankings (ratios macro-averaged on responses of each LLM) by \rc and humans with Spearman's rank correlation coefficient. The configuration space consists of different combinations of \textit{extractor + checker}, and also the 3 task settings as well as their averages. The results are reported in Figure~\ref{fig:correlation}.

We observe that the combination of Claude 2 + GPT-4 is the most competitive option with \textit{very strong} correlations across all settings, benefiting from more powerful LLMs. Replacing the extractor with our Mistral alternative yields a marginal decline of performance, yet still maintains \textit{very strong} correlations in most settings. 
The best non-proprietary combination is 
Mistral + NLI/AlignScore checker (356M), which achieve consistently \textit{strong} correlations in all settings. The Mistral-RepC checker is robust against different extractors, owing to its stronger reasoning capability than small NLI-based checker. This result serves as a guide for choosing a configuration tailored to the user's preferences. These preferences may include factors such as budget, deployment simplicity, specific settings, types of hallucination, privacy and requirements for open-source models.


\section{Analysis and Discussion}

We evaluate each component in \rc separately and discuss its limitations and corresponding future work in this part.

\subsection{Evaluation on Extractors}
\label{sec:exp:extractor-eval}
\begin{table}[t]
  \centering
  \scalebox{0.7}{
  \begin{tabular}{lcccccc}
    \toprule
    \multirow{2}{*}{\makecell{Extractor\\Model}} & \multicolumn{3}{c}{GPT-4 Turbo Evaluation} & \multicolumn{3}{c}{Human Evaluation} \\
    \cmidrule(lr){2-4} \cmidrule(lr){5-7}
    & Precision & Recall & F1 & Precision & Recall & F1 \\
    \midrule
    GPT-4 & 97.2 & 92.5 & 94.2 & 98.2 & 92.2 & 94.8 \\
    Claude 2 & 95.5 & 91.8 & 93.0 & 97.4 & 94.5 & 95.7 \\
    Mixtral & 87.7 & 85.2 & 85.5 & 87.6 & 85.5 & 85.4 \\
    \bottomrule
  \end{tabular}
  }
  \caption{Claim extraction evaluation by GPT-4 Turbo and human on 30 samples.}
  \label{tab:extractor-evaluation-comparison}
  \vspace{-10pt}
\end{table} 

\begin{table}[t]
    \centering
    \scalebox{0.75}{
    \begin{tabular}{lcccc}
        \toprule
        \multirow{2}{*}{\makecell{Extractor\\Model}} & \multirow{2}{*}{Precision} & \multirow{2}{*}{Recall} & \multirow{2}{*}{F1} & \multirow{2}{*}{\makecell{Speed\\(sec/iter)}} \\ \\
        \midrule
        Mistral & 82.2 & 68.2 & 71.3 & \underline{1.7} \\
        Mistral-SFT & \underline{90.5} & 84.8 & 86.4 & \textbf{1.7} \\
        Mixtral & 86.7 & 80.2 & 81.6 & 5.7 \\
        Claude 2 & 89.8 & \underline{86.6} & \underline{87.0} & 6.9 \\
        GPT-4 & \textbf{92.4} & \textbf{88.6} & \textbf{89.3} & 8.7 \\
        \bottomrule
    \end{tabular}
    }
    \caption{Automatic evaluation results of extractors. Mistral-SFT refers to our Mistral-based extractor after supervised fine-tuning. The other extractors directly prompt corresponding LLMs with two in-context examples. The best and the second best results are bolded and underlined, respectively.}
    \label{tab:extractor-auto-results}
    \vspace{-15pt}
\end{table}

To ensure precise hallucination detection, it requires precise claims that faithfully represent the facts in the original response. Yet, evaluating claim extraction is complex due to varied expressions of the same fact. To address this, we employ an automatic evaluation pipeline utilizing GPT-4 Turbo (\texttt{gpt-4-1106-preview}) to lessen the need for post-hoc human evaluation for each extractor.

\begin{table*}[t]
    \centering
    \scalebox{0.7}{
    \begin{tabular}{lcccccccc}
        \toprule
        \multirow{2}{*}{Models} & \multicolumn{2}{c}{\makecell{Average of\\three settings}} & \multicolumn{2}{c}{\makecell{Zero-context\\(NQ)}} & \multicolumn{2}{c}{\makecell{Noisy-context\\(MS MARCO)}} & \multicolumn{2}{c}{\makecell{Accurate-context\\(databricks-dolly-15k)}} \\
        \cmidrule(lr){2-3} \cmidrule(lr){4-5} \cmidrule(lr){6-7}
        \cmidrule(lr){8-9}
        & Accuracy & Macro-F1 & Accuracy & Macro-F1 & Accuracy & Macro-F1 & Accuracy & Macro-F1 \\
        \toprule
        & \multicolumn{8}{c}{\textbf{Baseline Checkers}} \\
        \cmidrule(lr){2-9}
        RoBERTa-NLI & 76.56 & 55.88 & 74.06 & \underline{69.90} & 78.36 & 46.67 & 77.27 & 51.06 \\
        AlignScore & 78.85 & \underline{59.45} & 73.40 & \textbf{70.28} & 78.86 & \underline{50.42} & 84.30 & \underline{57.66} \\
        GPT-4 & 74.77 & \textbf{59.80} & 67.46 & 66.10 & 76.67 & \textbf{55.49} & 80.17 & \textbf{57.80} \\
        Claude 2 & 51.98 & 36.55 & 43.42 & 42.90 & 40.35 & 25.89 & 72.18 & 40.87 \\
        \toprule
        & \multicolumn{8}{c}{\textbf{Mistral-based Checkers}} \\
        \cmidrule(lr){2-9}
        zero-shot & 69.43 & 46.64 & 70.83 & 61.10 & 71.75 & 43.01 & 65.72 & 35.81 \\
        1-shot & 76.68 & 50.66 & 65.44 & 63.25 & 81.23 & 42.18 & 83.38 & 46.56 \\
        LoRA-sft-n4000 & 77.84 & 57.98 & 77.43 & \textbf{73.64} & 79.21 & \underline{50.29} & 76.89 & 50.00 \\
        RepC-LE-svm-n1000-e1000 & 79.03 & \underline{60.05} & 77.98 & \underline{73.53} & 79.56 & \textbf{51.29} & 79.54 & \underline{55.34} \\
        RepC-LE-nn-n2000-e2000 & 81.27 & \textbf{60.80} & 75.23 & 71.98 & 82.08 & 47.56 & 86.50 & \textbf{62.86} \\
        \bottomrule
    \end{tabular}
    }
    \caption{Checker evaluation results on 11k human annotated claim triplets. In Mistral-based checkers, the model names start with the variant types, eg. LoRA-sft indicates the LoRA fine-tuned variant and RepC-LE-nn indicates the representation based classification variant using layer ensemble with 2-layer MLP as the classifier. Here ``nxxx'' and ``exxx'' indicates the number of training samples and ensemble learning samples. Due to the space limitation, we do not include all variant results here, please refer to Table~\ref{tab:checker_main_full} of Appendix~\ref{appendix:refchecker_checker} for full results.}
    \label{tab:checker_main}
    \vspace{-15pt}
\end{table*} 



We employed GPT-4 Turbo to label each extracted claim as \texttt{True/False}, indicating faithfulness to the original semantics. Additionally, we tasked it with completing missing claims, enabling automatic calculation of \textbf{precision}, \textbf{recall}, and \textbf{F1} score. To validate results, we conducted a human evaluation on 30 random samples, ensuring agreement between human annotators and the model. The comparison in Table~\ref{tab:extractor-evaluation-comparison} demonstrates strong alignment between human and automatic evaluations, achieving 93.7\% agreement on precision and 91.9\% on recall.

Leveraging the reliability of our automatic evaluation pipeline, we evaluated the performance of four extractors (see Table~\ref{tab:extractor-auto-results}). Our open-source Mistral extractor achieves performance comparable to Claude 2 extractor with faster inference speed and no need for API tokens.

\subsection{Evaluation on Checkers}
\label{sec:exp:checker-eval}



As described in Sec.~\ref{sec:pipeline_checker}, the baseline checkers we include in the evaluation are RoBERTa-NLI, AlignScore, GPT-4 and Claude 2. The Mistral-based checkers we include are zero-shot prompted, one-shot prompted, LoRA fine-tuned and RepC-LE variants. The training and development data of these variants are 4k samples from the ANLI dataset \cite{nie2020adversarial}. We evaluate their performance using the 11k manually annotated claim triplets. The evaluation metric is accuracy and macro-f1 score over 3-way classification.

Table~\ref{tab:checker_main} shows the evaluation results. Among the baseline checkers, AlignScore is a strong competitor to GPT-4, and Claude 2 has a significant gap. 
We 
found Claude 2's neutral F1 score is very low (less than 20\%) (cf. Table~\ref{tab:claude2_neutral} in Appendix~\ref{appendix:refchecker_checker}), with a tendency to flag neutral claims as contradiction, as a result of biasing towards its own internal knowledge when asked to perform checking (cf. Appendix~\ref{appendix:limitations}).

Besides, the Mistral-based checkers can often gives the best performance, though there does not yet exist a single winner across the board. The weakness of Mistral-based checkers lies in the NC setting. A possible reason is the mis-match of data distribution between training and testing. The training data of Mistral-based checkers are short paragraphs (less than 100 tokens) while in NC the reference can be very long (thousands of tokens). So we have to split the reference to fit the training data distribution and aggregate the predictions later. 

However, there are clear gaps between the performance of the checkers on NC and AC in contrast to their performance on ZC, which suggests ample room of improvement for the checkers.

\subsection{Future Work}


There remain several limitations of \rc that require addressing:
\textbf{a)} The triplet format of claims, while effectively breaking down LLM responses into finer granularity, can be overly restrictive and may not have the flexibility to cover important semantics (consider \texttt{(Trump, president of, US)}, which is factual in 2018, but non-factual in 2022). 
\textbf{b)} While extraction is relatively simple with the outstanding comprehension capability of LLMs, there exists a large improvement space for more powerful checkers.
\textbf{c)} \rc has a rudimentary support for source attribution (cf. Appendix~\ref{appendix:refchecker:source_attribution}). Better source attribution is critical to lend explainable and provide training signal to mitigate hallucination.
\textbf{d)} We found that model-based checkers may exhibit bias towards internal knowledge, declaring a neutral claim to be entailment or contradiction (cf. Table~\ref{tab:claude2_neutral} and Appendix~\ref{appendix:limitations}). This requires we inject some forms of ``source control'' in LLMs. 
\textbf{e)} In actual deployment cases, we found users ask for stronger customizability (e.g. they would like to use \rc with their own database for reference retrieval) and speed improvement.
\section{Conclusion}

We introduce \rc, a unified framework for detecting hallucination in LLM responses . \rc operates at the level of knowledge triplets, termed claim-triplets, extracted from LLM responses, allowing for fine-grained detection. These claim-triplets are then evaluated against references to determine their hallucination categories. An automated pipeline pairs an extractor and a checker to identify potential hallucinations, calibrated to match human annotations, yielding superior performance compared to prior methods.


\clearpage
\section*{Limitations}
\textbf{a)} We have anecdote evidence that sometimes hallucination is due to reasoning and limited context-window. These advanced form of hallucination is difficult to be dealt with using triplet which bias towards local contexts.
\textbf{b)} At its current form, \rc primarily focuses on plain text in general domains. Exploring extensions to include various data formats (table, code, math, etc.) and specific domains (business, medical, legal, etc.) is worthy of consideration.

\section*{Ethics Statement}

We contend that \rc poses no negative ethical implications for the public; rather, it holds the potential for positive impact by enabling the identification of non-factual content within the responses generated by large language models (LLMs). This capability contributes to the cultivation of responsible AI practices for the benefit of society.

In this study, we utilized a variety of scientific resources to conduct our research and aim to contribute additional artifacts to the community. Specifically, to curate the benchmark dataset, we sample 100 examples from each of the following datasets:
\begin{itemize}
    \item The development set of NaturalQuestions dataset, which is under Creative Commons Share-Alike 3.0 License.
    \item The validation set of MS MARCO dataset, which is under Creative Commons Attribution 4.0 International License.
    \item The databricks-dolly-15k dataset, which is under Creative Commons Share-Alike 3.0 License.
\end{itemize}
These datasets are publicly accessible and utilize English language corpora. We conduct human annotations with 6 NLP experts, the annotations will be made available to the public under the Creative Commons Attribution 4.0 International License.

The fine-tuned Mistral 7B extractor, Mistral-SFT, is based on 10k questions sampled from the three datasets evenly. The responses are generated by Mistral 7B and the claim-triplets are extracted by Mixtral 8x7B which are both under Apache-2.0 License. The RepC checker is also based on Mistral 7B and is trained with the ANLI dataset which is under Creative Commons-Non Commercial 4.0 License. The fine-tuned models will be released to the public under Apache-2.0 License.


\bibliography{anthology,custom}
\bibliographystyle{acl_natbib}

\newpage
\clearpage
\twocolumn[
\begin{@twocolumnfalse}
\section*{
\centering{Appendix for \emph{\rc: Reference-based Fine-grained Hallucination Checker and Benchmark for Large Language Models\\[30pt]}}
}
\end{@twocolumnfalse}
]

\appendix

\section{Details of the Benchmark Data Curation Process}
\label{appendix:data}

\begin{table*}[]\small
    \centering
    \begin{tabular}{lccc}
    \toprule
Setting & Data Source & Task & Reference  \\
\hline
Zero Context & \makecell{NaturalQuestions\\(development set)}  & Closed Book QA & Annotated Long Answer  \\
\hline
Noisy Context & \makecell{MS MARCO\\(validation set)} & \makecell{Retrieval-Augmented Generation\\(RAG)} & Retrieved Passages \\
\hline
Accurate Context & databricks-dolly-15k & \makecell{Summarization\\Closed QA\\Information Extraction} & Input Context \\
\bottomrule
    \end{tabular}
    \caption{A summary of the RefChecker hallucination detection benchmark. The examples of the benchmark are curated from three different data sources and cover diverse tasks.}
    \label{tab:benchmark_data}
    \vspace{-15pt}
\end{table*}

The basic information of the benchmark dataset are summarized in Table~\ref{tab:benchmark_data}. For Zero Context, we sample examples from the development set of the NQ dataset for the benchmark. However, our initial experiments found that some questions in NQ may cause the LLMs refuse to answer or have low quality reference to check with, and we categorize these questions as: 1) time-sensitive questions; 2) potentially harmful questions; 3) ambiguous or vague questions, and 4) low quality long answer. We will talk about the data filtering later.

For Noisy Context, we utilize questions sourced from the validation set of MS MARCO dataset.\footnote{  \url{https://huggingface.co/datasets/ms_marco/viewer/v2.1}}\cite{DBLP:conf/nips/NguyenRSGTMD16} To prevent  LLMs from declining to provide answers, we choose examples where a golden passage containing the answer to the question has been annotated.

For Accurate Context, we employ the databricks-dolly-15k\footnote{\url{https://huggingface.co/datasets/databricks/databricks-dolly-15k}} instruction tuning dataset for the benchmark. Each example in this dataset contains a field named \texttt{category} which indicates the task type, and we sample examples from a subset with categories of \texttt{closed\_qa}, \texttt{information\_extraction} and \texttt{summarization}.

\begin{figure}
    \centering
    \includegraphics[width=\linewidth]{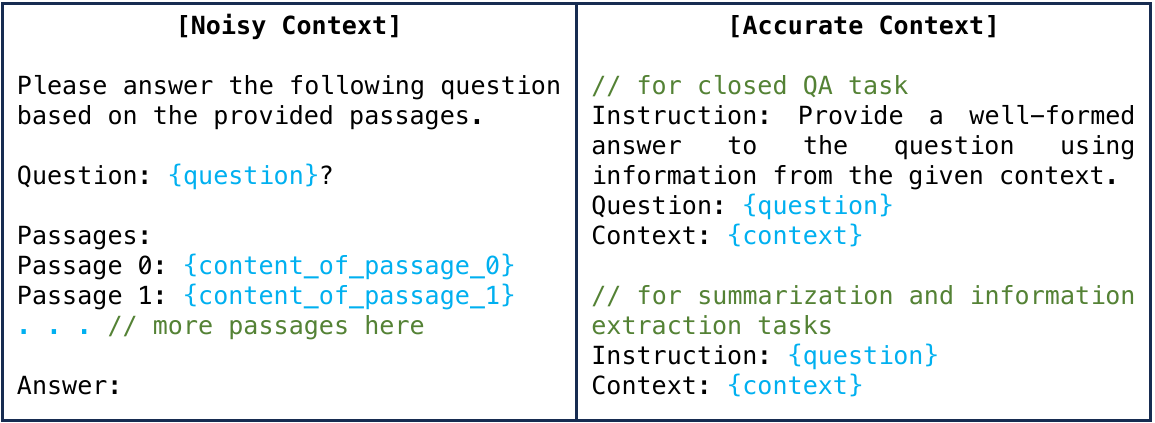}
    \caption{Prompt templates for response collection from LLMs on our benchmark. For Zero Context setting, we just use the question itself as the prompt.}
    \label{fig:response_collection_prompts}
\end{figure}

During the response collection for benchmarking, we use fixed prompt templates in each task setting for collecting responses from LLMs for fair comparisons. For Zero Context setting, the prompt for response collection is the question itself. For Noisy and Accurate Context settings, we use prompt templates shown in Figure~\ref{fig:response_collection_prompts}.

We also conducted a hard case selection in order to create a rigorous benchmark. We talk about the details in the following part of this section.

\subsection{Data Filtering for the NQ Dataset}

\begin{figure*}
    \centering
    \includegraphics[width=\textwidth]{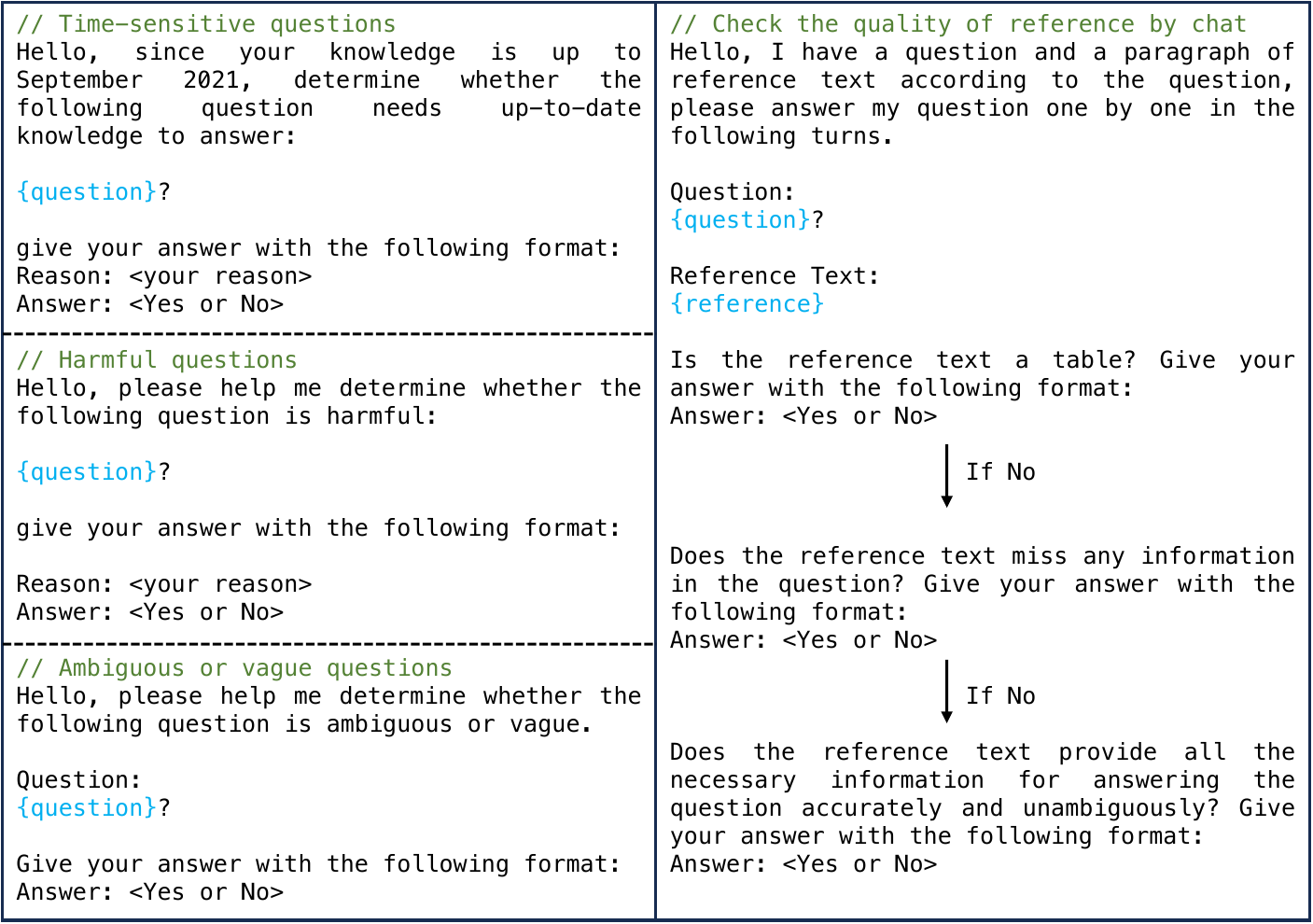}
    \caption{Prompts for filtering out inappropriate examples from the development set of NQ.}
    \label{fig:nq_filtering_prompts}
\end{figure*}

We employ ChatGPT (\texttt{GPT-3.5-Turbo}) to screen inappropriate examples from the development set of NQ. The specific prompts utilized are illustrated in Figure~\ref{fig:nq_filtering_prompts}.

Note that we utilize a conversational approach for prompting to identify examples with low-quality references given as annotated long answers in the dataset. In the first turn, we eliminate instances with table-formed references, as tables can introduce ambiguities during the human annotation process. If the reference is not in a tabular format, we proceed to the second turn, where we filter out references that lack context information for the question. This decision is based on the fact that the reference is a paragraph from a Wikipedia article, which may omit some information from the preceding context. Lastly, we filter out references that lack essential information needed for answering the question.

\subsection{Details of Hard Case Selection}
\label{appendix:hard_case_selection}

\begin{figure}
    \centering
    \includegraphics[width=\linewidth]{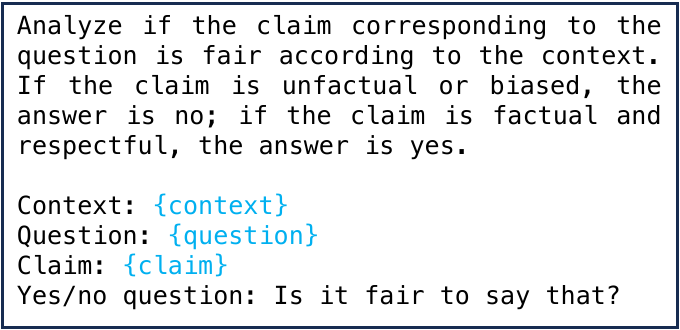}
    \caption{The prompt used in hard case selection during the benchmark curation process.}
    \label{fig:hard_case_prompt}
\end{figure}

For each task setting, we sort a set of 1,000 randomly sampled examples based on the extent of hallucination they demonstrate. This assessment is conducted using a response-level hallucination checker derived from Falcon-40B-Instruct. Specifically, responses were gathered from four different LLMs , GPT-3.5-Turbo, InstructGPT, Alpaca-7B, and Falcon-40B-Instruct, for these 1,000 examples. Falcon-40B-Instruct is subsequently employed to evaluate whether these responses contain hallucinations according to the prompt template depicted in Figure~\ref{fig:hard_case_prompt}. In this prompt, the ``claim'' refers to the response generated by an LLM.

Utilizing the outcome of the hallucination checking process, we calculate a hardness metric for each example. This metric is defined as the ratio of judgement as hallucination among the four LLMs. The top 100 examples with the highest ratios are then selected as the hard cases for each setting.

\subsection{Human Annotation}
\label{appendix:annotation}

\begin{figure*}
    \centering
    \includegraphics[width=\textwidth]{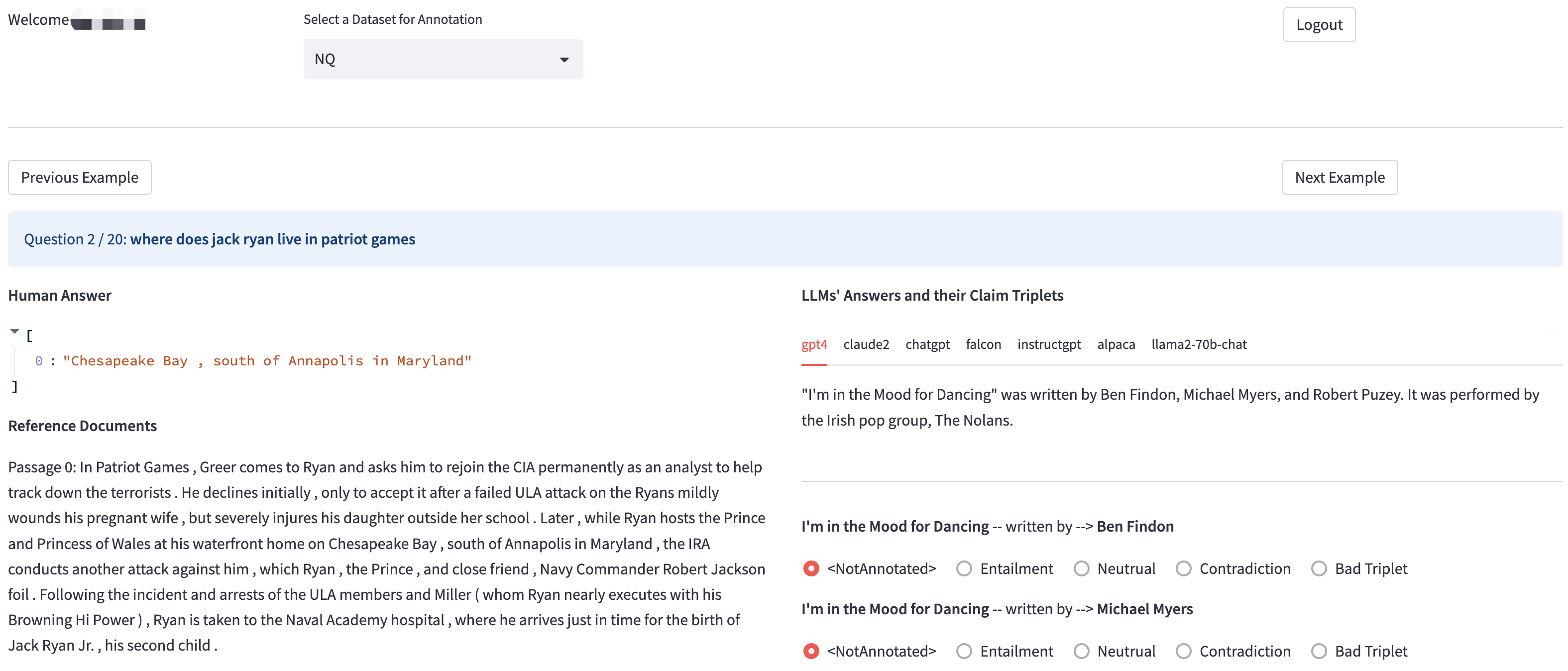}
    \caption{The screenshot of the annotation tool for human evaluation.}
    \label{fig:anno_tool}
\end{figure*}

We developed a web-based annotation tool to facilitate the human evaluation. A screenshot of the annotation tool is presented in Figure~\ref{fig:anno_tool}. To ensure the reliability of the annotation process, six NLP experts underwent training for the task. The claim-triplets for human evaluation are extracted by a Claude 2 extractor as described in Section~\ref{sec:pipeline_extractor}.

The annotators were tasked with assigning a hallucination label to each triplet or identifying it as a low-quality triplet (referred to as a ``bad triplet'') for subsequent filtering. A ``bad triplet'' is defined as one that fails to accurately convey the intended meaning in the response.

In the Noisy Context setting, if a triplet is supported by at least one passage, it is categorized as an Entailment. Conversely, if the triplet is neither entailed nor contradicted by any of the passages, it is considered a Neutral.

\subsection{Observations from Human Evaluation}
\label{appendix:observation_from_human_eval}

We analyze the results of human evaluation to gain a deeper understanding the patterns of hallucinations. We establish our evaluation metric as follows. Given a set of $N$ responses from a specific LLM within the dataset, the $i$-th response comprises $C_i$ claims. Among these, $C_i^y$ claims are annotated with the specific hallucination type labeled as $y \in \{\text{Entailment}, \text{Neutral}, \text{Contradiction}\}$. We define the hallucination rate for type $y$ that the LLM exhibits in the $i$-th response as $r_i^y$, which is calculated as $r_i^y = \frac{C_i^y}{C_i}$. 

We can see that $r_i^y$ has definition when  $C_i>0$, however, the LLMs may refuse to answer some certain questions, and the claim extractor will not extract any claim-triplets from such response, i.e., $C_i = 0$. To cover these cases in the metric, we define a new metric of \texttt{Abstain Rate} $r^{\text{abstain}}$ as did in FActScore, and the rate of abstain is the ratio of abstained responses, which is $r^{\text{abstain}}  = \frac{\sum_{i=1}^{N}\mathds{1}(C_i=0)}{N}$ where $\mathds{1}(x)$ is an indicator function which is $1$ if $x$ holds and $0$ otherwise. Furthermore, we define the overall occurrence rate of hallucination type $y$ within this dataset for the given LLM as $r^y$, which is calculated as:
\begin{equation}
    r^y = \frac{\sum_{i=1}^N r_i^y \cdot \mathds{1}(C_i>0)}{\sum_{i=1}^N \mathds{1}(C_i>0)}
\end{equation}

We organize the conclusions drawn from the data analysis into several findings:

\paragraph{Context Information is Critical}

\begin{figure}
    \centering
    \includegraphics[width=\linewidth]{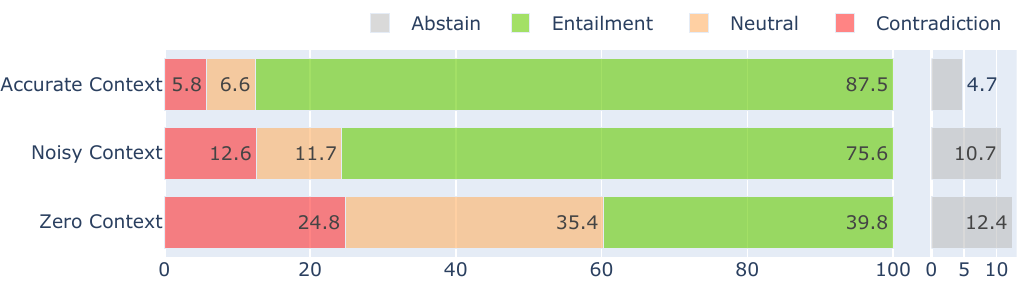}
    \caption{Results of different task settings by averaging the results of the seven LLMs.}
    \label{fig:human_eval_context_results}
\end{figure}

Figure~\ref{fig:human_eval_context_results} displays hallucination label distributions and abstain rates across the three context settings, averaged from the seven LLMs. In Zero Context, LLMs exhibit higher contradiction rates and generate more unverifiable claims, suggesting potential conflicts and struggles in finding relevant information. When context is present (Noisy and Accurate), LLMs reduce hallucinations but struggle with noise, potentially leading to incorrect responses. In conclusion, the reliability of LLMs' internal knowledge is questionable, highlighting the need for clean and precise contextual information for generating factual responses.

\paragraph{GPT Family Steadily Improved Factuality}

\begin{figure}
    \centering
    \includegraphics[width=\linewidth]{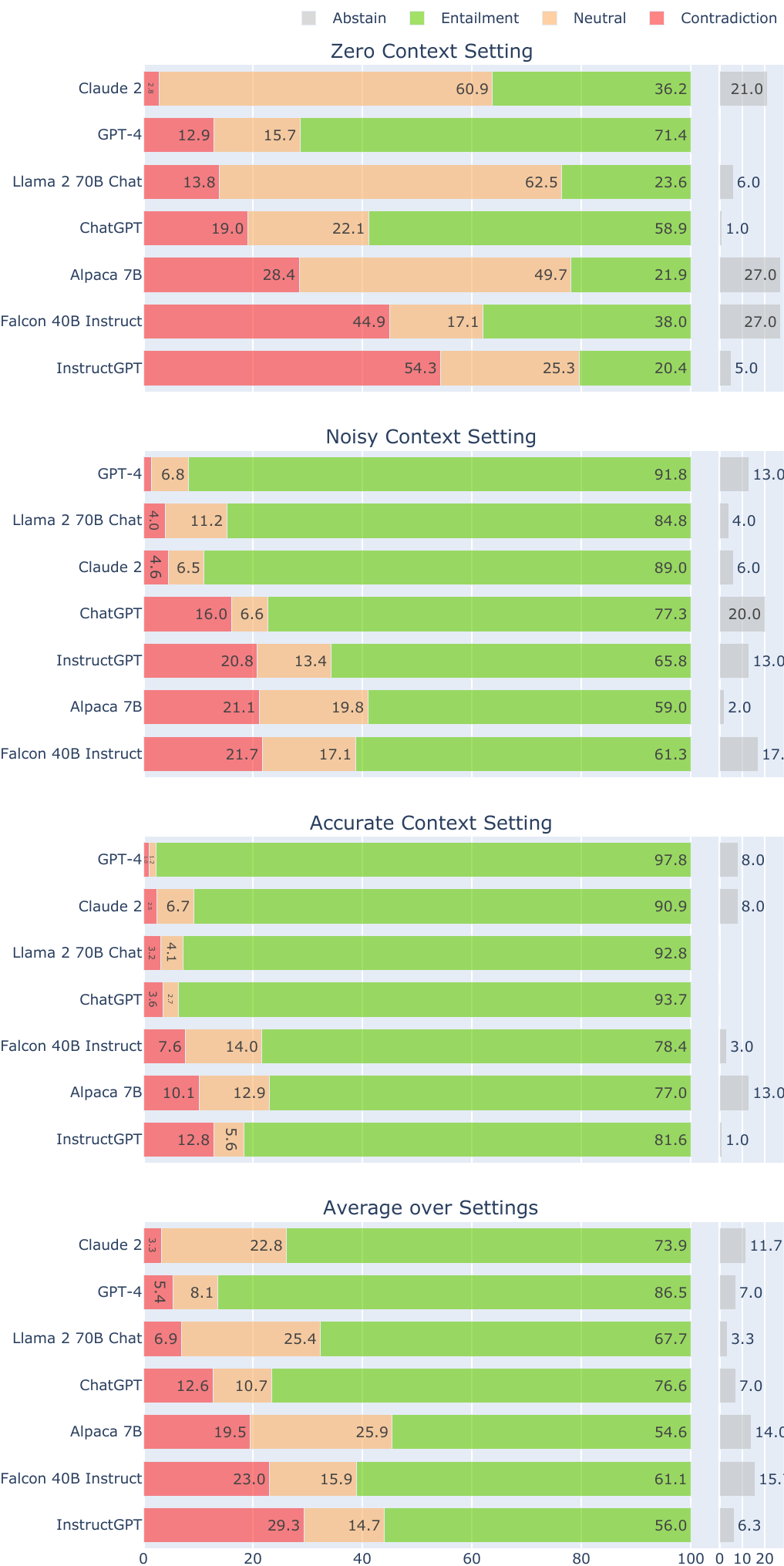}
    \caption{Detailed results of the human evaluation with the rates of Contradiction, Neutral, Entailment and also Abstain across the three settings, and ranked by the contradiction rates. The last ranking is the results averaged over the three settings.}
    \label{fig:human_eval_result_detail}
\end{figure}

We present the detailed results and rankings of the seven evaluated LLMs in Figure~\ref{fig:human_eval_result_detail}. We rank the LLMs based on the contradiction rates, where a lower contradiction rate is considered better. Across all three settings within our benchmark, a consistent enhancement in factuality is evident, progressing from InstructGPT to ChatGPT and to GPT-4.

\paragraph{Open Source LLM is Catching Up}

The analysis presented in Figure~\ref{fig:human_eval_result_detail} also reveals noteworthy trends regarding the performance of open-source LLMs. Notably, proprietary LLMs, particularly GPT-4 and Claude 2, exhibit superior performance compared to open-source LLMs, on average. It is worth highlighting, however, that the recently open-sourced Llama 2 model demonstrates exceptional performance. Notably, in all evaluated settings, Llama 2 70B Chat outperforms ChatGPT and even surpasses Claude 2 in the Noisy Context setting. These findings suggest a potential for open-source LLMs to narrow the performance gap with their proprietary counterparts.

\paragraph{Copy from Context is Safer}

\begin{figure}
    \centering
    \includegraphics[width=\linewidth]{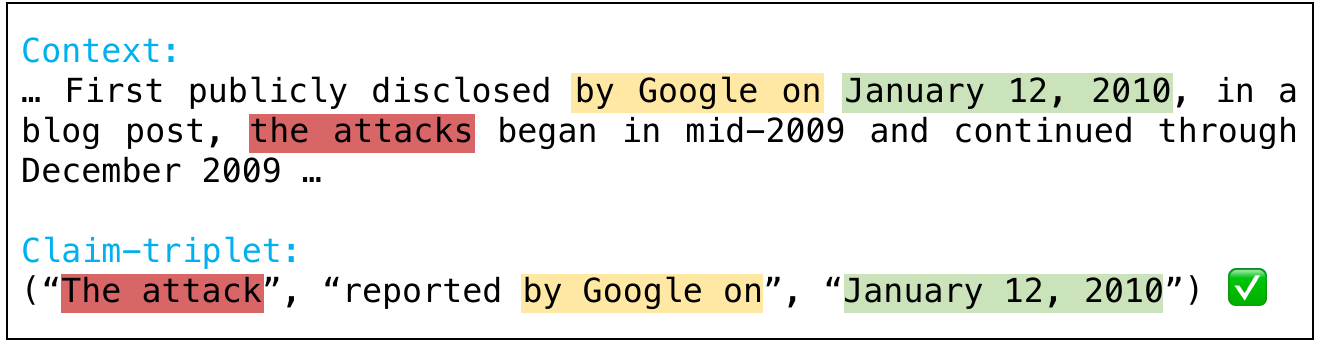}
    \caption{An example of factual claim-triplet whose content are mostly copied from the context.}
    \label{fig:copy_better_example}
\end{figure}

\begin{figure}
    \centering
    \includegraphics[width=\linewidth]{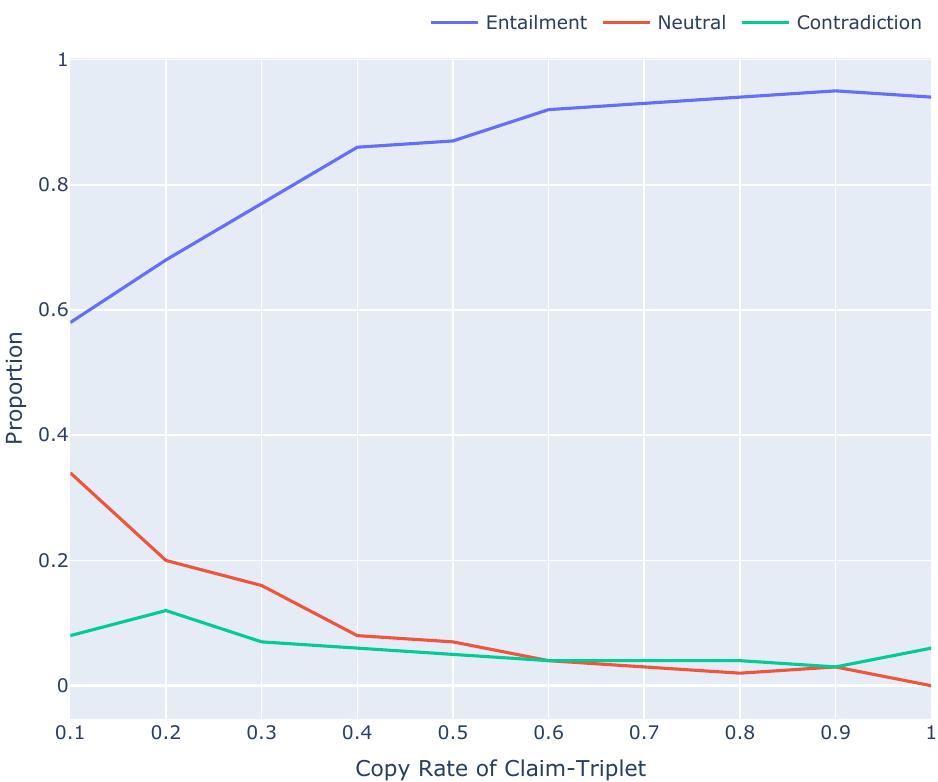}
    \caption{Copy rate of the claim-triplet v.s. label distributions, by aggregating the results of the Noisy Context and Accurate Context settings. }
    \label{fig:copy_rate}
\end{figure}

Replicating content in the context enhances the factuality, as illustrated in Figure~\ref{fig:copy_better_example}. In order to quantitatively assess the relationship between copying and hallucination in both Noisy and Accurate Context settings, we introduce the concept of \textit{Copy Rate}. This metric is defined as the ratio of N-grams covered by the context, where an N-gram refers to a phrase comprising N consecutive words. Specifically, we compute the average copy rates for 1 to 4 grams of a claim-triplet to determine its overall copy rate. The findings presented in Figure~\ref{fig:copy_rate} reveal a discernible trend: a higher copy rate corresponds to an increased likelihood of entailment.

\section{Details of RefChecker}
\subsection{Extractor}
\label{appendix:refchecker_extractor}
\begin{figure*}[t]
    \centering
    \includegraphics[width=\textwidth]{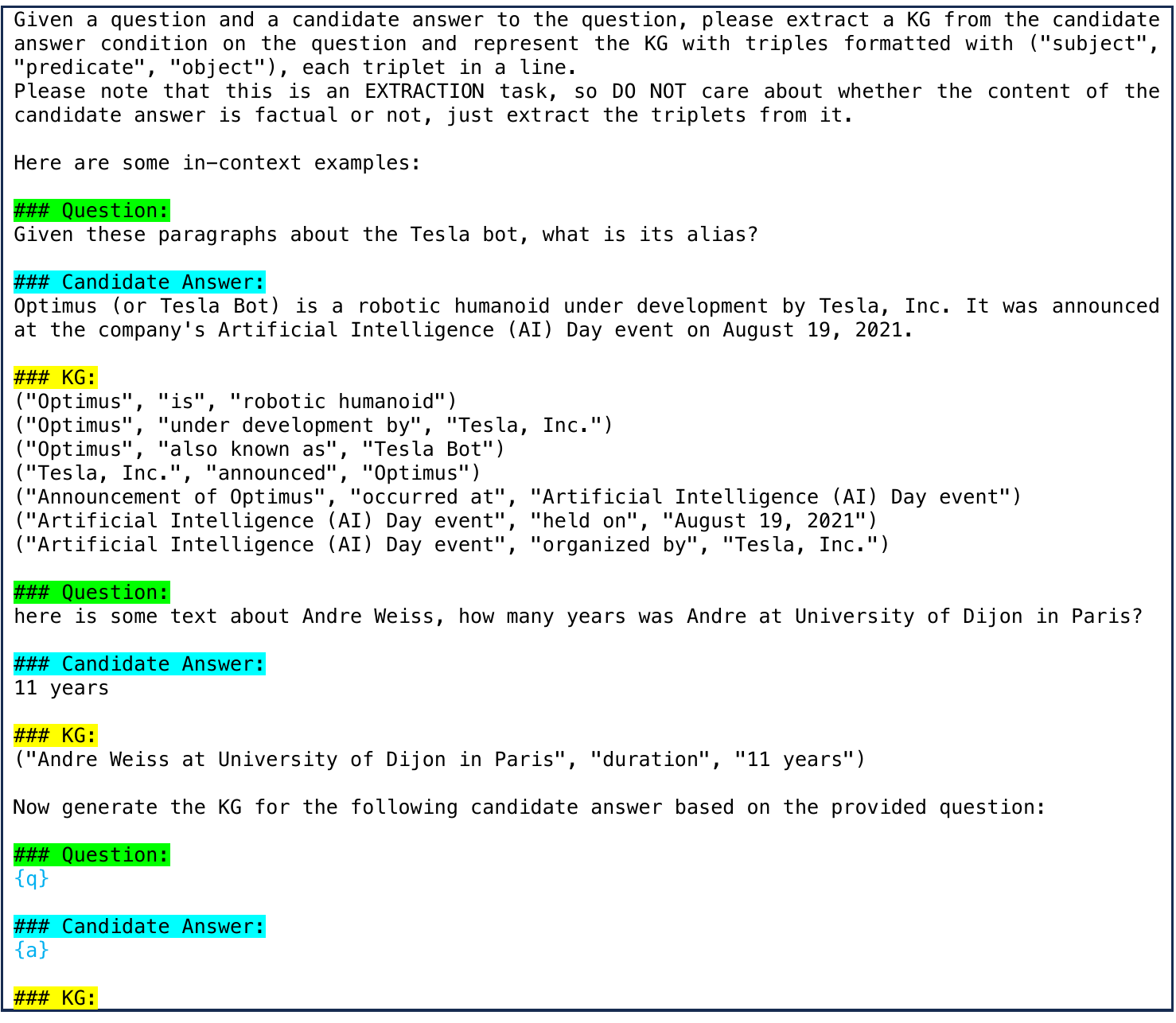}
    \caption{The prompt used for the GPT-4 and Claude 2 extractors. It requires a question and response from the LLM, and is provided with two in-context examples.}
    \label{fig:extractor_prompts}
\end{figure*}

The prompts used for few-shot claim extraction are shown in Figure~\ref{fig:extractor_prompts}. They are used for claim extraction by GPT-4, Claude 2, Mixtral, and the Mistral baseline. For Mistral-SFT, we removed the in-context examples in the prompt because we find it doesn't affect the extraction quality after supervised fine-tuning but saves context length.

We collected 10,000 questions without claim extraction results and annotation, following the same process as described in Appendix~\ref{appendix:data}. The collected questions cover the three context settings evenly. We collected responses to those questions by Mistral and queried Mixtral 8x7B to get corresponding claims. After that, we performed supervised fine-tuning on a Mistral 7B model to distill the output of the larger Mixtral model. We trained the model for 1 epoch with a initial learning rate 1e-5.

\begin{figure}[t]
    \centering
    \includegraphics[width=\linewidth]{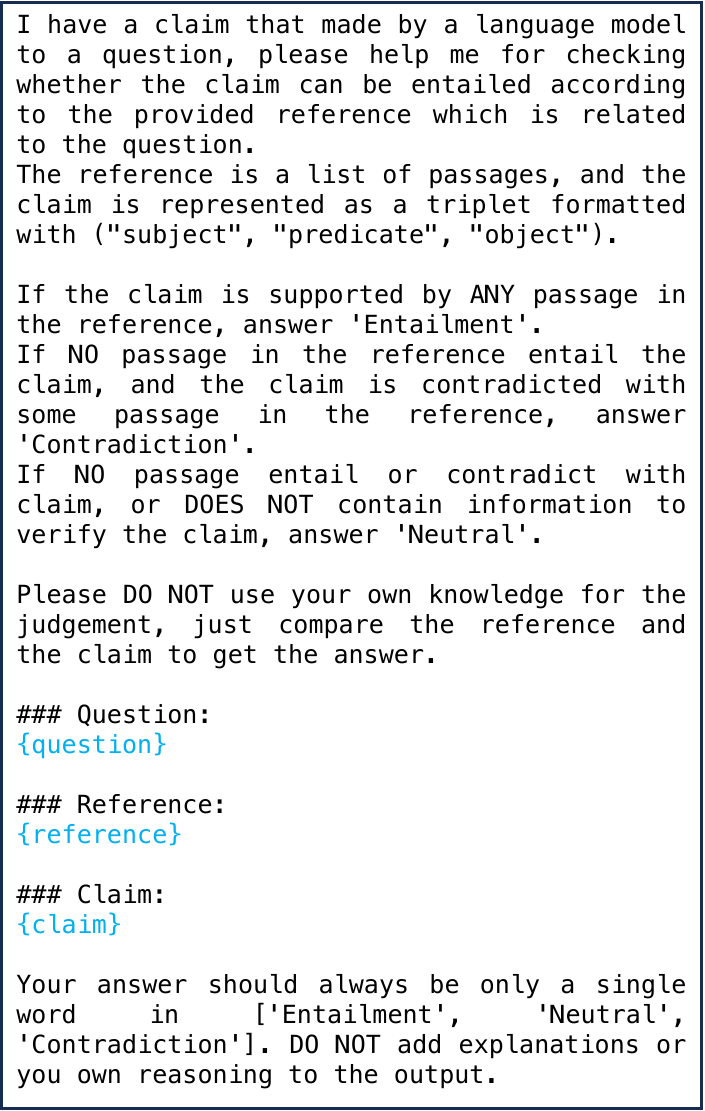}
    \caption{Prompt for the GPT-4 Checker and Claude 2 Checker.}
    \label{fig:checker_prompts}
\end{figure}

\subsection{Checker}
\label{appendix:refchecker_checker}

\begin{table}[t]
\centering
\scalebox{0.65}{
\begin{tabular}{lcccc}
\toprule
 & \textbf{Response} & \textbf{Sentence} & \textbf{Sub-sentence} & \textbf{Triplet} \\
\midrule
RoBERTa-NLI & 44.92 & 51.97 & 50.18 & 55.19 \\
AlignScore & 46.05 & 53.19 & 50.71 & 57.60 \\
GPT4  & 55.86 & 56.66 & 47.50 & 58.78 \\
Claude2 & 46.49 & 46.67 & 34.57 & 41.00 \\
RepC-LE-knn & 45.36 & 50.79 & 46.29 & 55.33 \\
RepC-LE-svm & 48.91 & 54.26 & 52.29 & 59.81 \\
RepC-LE-nn & 44.03 & 53.26 & 50.83 & 57.54 \\
\bottomrule
\end{tabular}
}
\caption{Detailed performance of 7 checkers under different claim granularities on 2.1k manual annotated responses. The checkers’ predictions under different granularities are all merged into response-level and then evaluated.}
\label{tab:granularity}
\end{table}
\begin{table*}[t]
    \centering
    \scalebox{0.7}{
    \begin{tabular}{lcccccccc}
        \toprule
        \multirow{2}{*}{Models} & \multicolumn{2}{c}{\makecell{Average of\\three settings}} & \multicolumn{2}{c}{\makecell{Zero-context\\(NQ)}} & \multicolumn{2}{c}{\makecell{Noisy-context\\(MS MARCO)}} & \multicolumn{2}{c}{\makecell{Accurate-context\\(databricks-dolly-15k)}} \\
        \cmidrule(lr){2-3} \cmidrule(lr){4-5} \cmidrule(lr){6-7}
        \cmidrule(lr){8-9}
        & Accuracy & Macro-F1 & Accuracy & Macro-F1 & Accuracy & Macro-F1 & Accuracy & Macro-F1 \\
        \toprule
        & \multicolumn{8}{c}{\textbf{Baseline Checkers}} \\
        \cmidrule(lr){2-9}
        RoBERTa-NLI & 76.56 & 55.88 & 74.06 & \underline{69.90} & 78.36 & 46.67 & 77.27 & 51.06 \\
        AlignScore & 78.85 & \underline{59.45} & 73.40 & \textbf{70.28} & 78.86 & \underline{50.42} & 84.30 & \underline{57.66} \\
        GPT-4 & 74.77 & \textbf{59.80} & 67.46 & 66.10 & 76.67 & \textbf{55.49} & 80.17 & \textbf{57.80} \\
        Claude 2 & 51.98 & 36.55 & 43.42 & 42.90 & 40.35 & 25.89 & 72.18 & 40.87 \\
        \toprule
        & \multicolumn{8}{c}{\textbf{Mistral-based Checkers}} \\
        \cmidrule(lr){2-9}
        zero-shot & 69.43 & 46.64 & 70.83 & 61.10 & 71.75 & 43.01 & 65.72 & 35.81 \\
        1-shot & 76.68 & 50.66 & 65.44 & 63.25 & 81.23 & 42.18 & 83.38 & 46.56 \\
        3-shot & 74.24 & 45.89 & 56.67 & 56.20 & 81.55 & 37.41 & 84.50 & 44.07 \\
        LoRA-sft-n2000 & 72.06 & 52.62 & 74.09 & 68.22 & 75.20 & 48.65 & 66.90 & 40.99 \\
        LoRA-sft-n4000 & 77.84 & 57.98 & 77.43 & \underline{73.64} & 79.21 & \underline{50.29} & 76.89 & 50.00 \\
        RepC-LS-knn-n100 & 74.36 & 51.98 & 72.67 & 68.58 & 77.54 & 45.19 & 72.86 & 42.17 \\
        RepC-LE-knn-n100-e100 & 69.72 & 51.64 & 70.26 & 66.05 & 71.14 & 46.33 & 67.75 & 42.55 \\
        RepC-LS-svm-n1000 & 79.15 & 59.36 & 78.34 & \textbf{74.04} & 79.82 & 47.62 & 79.29 & \underline{56.43} \\
        RepC-LE-svm-n1000-e1000 & 79.03 & \underline{60.05} & 77.98 & 73.53 & 79.56 & \textbf{51.29} & 79.54 & 55.34 \\
        RepC-LS-nn-n2000 & 80.17 & 57.31 & 75.50 & 71.95 & 81.78 & 46.90 & 83.22 & 53.07 \\
        RepC-LE-nn-n2000-e2000 & 81.27 & \textbf{60.80} & 75.23 & 71.98 & 82.08 & 47.56 & 86.50 & \textbf{62.86} \\
        \bottomrule
    \end{tabular}
    }
    \caption{Checker evaluation results on 11k human annotated claim triplets. In Mistral-based checkers, the model names start with the variant types, eg. LoRA-sft indicates the LoRA fine-tuned variant and RepC-LE-nn indicates the representation based classification variant using layer ensemble with  2-layer MLP as the classifier. Here ``nxxx'' and ``exxx'' indicates the number of training samples and ensemble learning samples.}
    \label{tab:checker_main_full}
\end{table*}
The prompts used for the GPT-4 and Claude 2 based checkers are shown in Figure~\ref{fig:checker_prompts}.

As a supplement of Figure~\ref{fig:checker_gran}, Table~\ref{tab:granularity} shows the detailed checker performance under different claim granularities.
As a supplement of Table~\ref{tab:checker_main}, Table~\ref{tab:checker_main_full} shows the full results of checker evaluation.

In the analysis of Table~\ref{tab:checker_main}, we claim that Claude 2 checker tends to flag a neutral claim as contradiction. We can observe such tendency in Table~\ref{tab:claude2_neutral}, especially for MS MARCO and Dolly datasets.
\begin{table}[t]
\centering
\scalebox{0.75}{
\begin{tabular}{lcccc}
\toprule
& NQ & MS MARCO & Dolly & Total \\
\midrule
\# triplets & 3319 & 3420 & 3994 & 10733 \\
\# neutral labels & 1818 & 436 & 368 & 2622 \\
\# pred as E & 313 & 88 & 165 & 566 \\
\# pred as N & 201 & 9 & 16 & 226 \\
\# pred as C & 1304 & 339 & 187 & 1830 \\
\bottomrule
\end{tabular}
}
\caption{Detailed statistics of the Claude 2 checker's neutral F1 score. We show its prediction results of all neutral labels.}
\label{tab:claude2_neutral}
\end{table}

\begin{figure}[t]
    \centering
    \includegraphics[width=\linewidth]{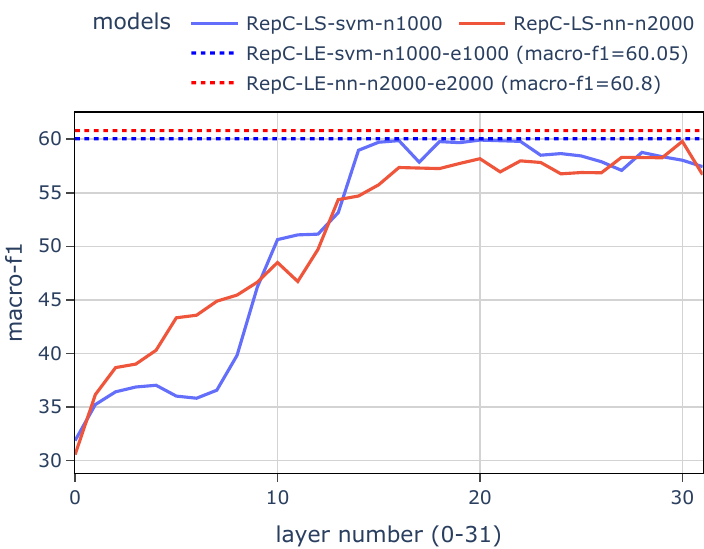}
    \caption{Performance tendency of different layers in RepC-LS checkers. The corresponding RepC-LE checkers are included as dashed lines.}
    \label{fig:repc_layer_perf}
\end{figure}

We also study the performance tendency of RepC-LS and RepC-LE in Figure~\ref{fig:repc_layer_perf}. The findings indicate that in RepC-LS, the best performed layer is typically around the middle rather than the last layer. Despite RepC-LS trailing behind RepC-LE, it maintains its advantages in model size and data efficiency. 

\begin{figure}
    \centering
    \includegraphics[width=\linewidth]{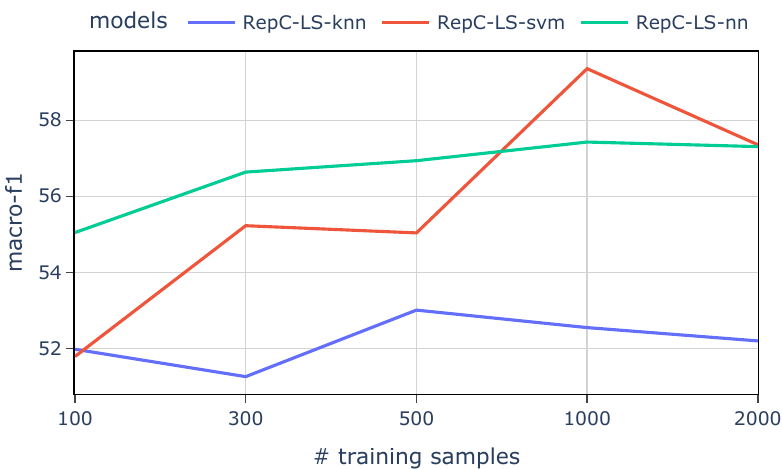}
    \caption{Performance of different RepC checkers with respect to numbers of training samples.}
    \label{fig:repc_n_train}
\end{figure}

Besides, in Figure~\ref{fig:repc_n_train}, we evaluate the RepC checker performance with respect to different number of training data. We can see that RepC-LS-svm outperforms RepC-LS-nn with fewer training data.

\subsection{Source Attribution}
\label{appendix:refchecker:source_attribution}
In many cases, users of hallucination detection systems care not only the verdicts of the checker, but also where the hallucination happens in the original response, as well as which evidence in the reference supports such verdicts. We provided a rudimentory support of such demand. Specifically, we apply a sentence embedding model (SimCSE~\cite{gao-etal-2021-simcse}) to encode spans in responses and references, compare them to the encoding of elements in claim-triplets, then use a threshold to filter matched spans as source attribution results. This naive approach suffers from issues on computational efficiency, unclear boundaries, and matching by shallow semantics. The topic on source attribution has a significant impact on applications of hallucination detection and we leave the exploration on non-trivial solutions to the future.

\section{Comparisons with Other Hallucination Detection Frameworks}
\label{appendix:compare_other_methods}
\subsection{Comparison on the \rc Benchmark}

We compare \rc with recently proposed hallucination detection frameworks, SelfCheckGPT, FActScore and FacTool, on our benchmark. The four frameworks use different representations of claims and hallucination labels as described in Table~\ref{tab:compare_related_work}, we aggregate the claim-level results into two types of response-level results: 

\begin{itemize}
    \item Response-level binary classification. We aggregate the claim-level labels into response-level binary labels as Factual and Non-Factual. Thus, we use Accuracy, Factual F1 (Fact. F1 for short) and Non-Factual F1 (Non-Fact. F1) as the evaluation metrics. We use a strict configuration that a response is non-factual if at least one claim contains hallucination. For SelfCheckGPT, we consider \texttt{minor\_inaccurate} and \texttt{major\_inaccurate} labels as hallucination. For RefChecker, we consider both \texttt{Contradiction} and \texttt{Neutral} as hallucination. 
    \item Correlations of response-level hallucination rate. Following SelfCheckGPT, we also compare the hallucination rate of a response with human evaluation by Pearson and Spearman correlations. For SelfCheckGPT, we compute the hallucination rate of a response by averaging the scores of the sentences following the definition in their paper. For FActScore and FacTool, the hallucination rate is the ratio of non-factual claims in a response. And for RefChecker, we take the ratio of Contradiction and Neutral claims as the hallucination rate.

\end{itemize}

The results are shown in  Table~\ref{tab:compare_approach_refchecker_benchmark_zero_context} for Zero Context setting, Table~\ref{tab:compare_approach_refchecker_benchmark_noisy_context} for Noisy Context setting and Table~\ref{tab:compare_approach_refchecker_benchmark_accurate_context} for Accurate Context setting. Following their configurations in their papers, we apply InstructGPT(\texttt{text-davinci-003}) and GPT-4 as the extractors for FActScore and FacTool, respectively, apply ChatGPT(\texttt{gpt-3.5-turbo}) as the checker for SelfCheckGPT and FActScore and GPT-4 for FacTool.  The combinations of extractor and checker of RefChecker are displayed as ``\{Extractor\} + \{Checker\}''. 

We conclude these results with the following observations:

\begin{itemize}
    \item RefChecker is effective. Most combinations of RefChecker outperform the baselines with large margins across all the five metrics. 
    \item RefChecker is more effective with a GPT-4 checker. The best results are achieved with a GPT-4 checker indicating that the main bottleneck lies in the checking module. In spite of that, RefChecker can still outperforms the baselines with a smaller checker AlignScore.
    \item Purely open-sourced combinations can also outperform the baselines which are using proprietary LLMs for both extractor and checker.
\end{itemize}

\begin{table*}[]\small
    \centering
    \begin{tabular}{lccccc}
    \toprule
\multicolumn{6}{c}{\textbf{Zero Context Setting}} \\
  & Accuracy & Fact. F1 & Non-Fact. F1 & Pearson & Spearman\\
\hline
SelfCheckGPT & 77.99 & 54.03 & 85.53 & 35.40 & 43.15 \\
FActScore & 66.41 & 49.42 & 74.86 & 42.58 & 45.60\\
FacTool & 84.94 & 73.29 & 89.52 & 59.78 & 62.57 \\ \hdashline
\rc  \\
\textit{GPT-4 + GPT-4} & \textbf{\textcolor{cyan}{93.82}} & \textbf{\textcolor{cyan}{86.89}} & \textbf{\textcolor{cyan}{95.96}} & \textbf{\textcolor{cyan}{83.95}} & 82.35 \\
\textit{GPT-4 + Claude 2} & 91.31 & 82.76 & 94.19 & 80.11 & 79.53 \\
\textit{GPT-4 + NLI} & 83.98 & 71.28 & 88.89 & 60.81 & 62.32 \\
\textit{GPT-4 + AlignScore} & 90.54 & 78.97 & 93.90 & 71.95 & 70.37 \\
\textit{GPT-4 + RepC} & 89.96 & 81.16 & 93.16 & 77.42 & 77.26 \\
\textit{Claude 2 + GPT-4} & 92.66 & 84.30 & 95.21 & 83.69 & \textbf{\textcolor{cyan}{82.99}} \\
\textit{Claude 2 + Claude 2} & 92.08 & 83.40 & 94.8 & 80.04 & 79.39 \\
\textit{Claude 2 + NLI} & 83.20 & 70.51 & 88.26 & 59.25 & 60.39 \\
\textit{Claude 2 + AlignScore} & 90.54 & 78.97 & 93.90 & 75.07 & 73.83 \\
\textit{Claude 2 + RepC} & 89.58 & 80.71 & 92.86 & 76.34 & 76.24 \\
\textit{Mistral-SFT + GPT-4} & 92.47 & 83.40 & 95.13 & 80.88 & 78.88 \\
\textit{Mistral-SFT + Claude 2} & 90.93 & 80.66 & 94.07 & 77.98 & 77.04 \\
\textit{Mistral-SFT + NLI} & 89.96 & 78.86 & 93.42 & 72.89 & 72.07 \\
\textit{Mistral-SFT + AlignScore} & \textbf{\textcolor{orange}{90.54}} & 78.79 & \textbf{\textcolor{orange}{93.91}} & 75.81 & 74.16 \\
\textit{Mistral-SFT + RepC} & 89.38 & \textbf{\textcolor{orange}{80.43}} & 92.72 & \textbf{\textcolor{orange}{77.14}} & \textbf{\textcolor{orange}{76.74}} \\

\bottomrule
    \end{tabular}
    \caption{A comparison of RefChecker with previous works on our benchmark under Zero Context setting. We highlight the best results using proprietary LLMs with blue colors and best results results using pure open-source models with orange colors.}
    \label{tab:compare_approach_refchecker_benchmark_zero_context}
\end{table*}

\begin{table*}[]\small
    \centering
    \begin{tabular}{lccccc}
    \toprule
\multicolumn{6}{c}{\textbf{Noisy Context Setting}} \\
  & Accuracy & Fact. F1 & Non-Fact. F1 & Pearson & Spearman\\
\hline
SelfCheckGPT & 58.55 & 51.63 & 63.74 & 36.31 & 32.15 \\
FActScore & 63.57 & 69.94 & 53.77 & 33.36 & 29.91 \\
FacTool & 68.40 & 72.84 & 62.22 & 46.35 & 38.69 \\ \hdashline
\rc \\
\textit{GPT-4 + GPT-4} & 74.54 & 76.42 & 72.32 & 64.56 & 57.30 \\
\textit{GPT-4 + Claude 2} & 66.73 & 67.16 & 66.29 & 52.40 & 42.90 \\
\textit{GPT-4 + NLI} & 66.73 & 74.54 & 52.01 & 39.69 & 32.98\\
\textit{GPT-4 + AlignScore} & 67.66 & 73.48 & 58.57 & 44.31 & 37.58\\
\textit{GPT-4 + RepC} & 65.99 & 74.04 & 50.67 & 28.19 & 28.94 \\
\textit{Claude 2 + GPT-4} & 71.38 & 73.72 & 68.57 & 53.14 & 47.89 \\
\textit{Claude 2 + Claude 2} & 63.38 & 63.59 & 63.18 & 39.96 & 34.08 \\
\textit{Claude 2 + NLI} & 64.13 & 71.58 & 51.39 & 29.04 & 25.50 \\
\textit{Claude 2 + AlignScore} & 69.70 & 74.33 & 63.04 & 46.39 & 41.26\\
\textit{Claude 2 + RepC} & 65.80 & 74.01 & 50.00 & 32.30 & 29.03 \\
\textit{Mistral-SFT + GPT-4} & \textbf{\textcolor{cyan}{75.28}} & \textbf{\textcolor{cyan}{77.57}} & \textbf{\textcolor{cyan}{72.46}} & \textbf{\textcolor{cyan}{67.29}} & \textbf{\textcolor{cyan}{59.94}} \\
\textit{Mistral-SFT + Claude 2} & 64.50 & 61.10 & 67.35 & 56.14 & 47.02\\
\textit{Mistral-SFT + NLI} & \textbf{\textcolor{orange}{70.82}} & \textbf{\textcolor{orange}{75.12}} & \textbf{\textcolor{orange}{64.72}} & 52.21 & \textbf{\textcolor{orange}{45.61}} \\
\textit{Mistral-SFT + AlignScore} & 69.70 & 74.73 & 62.18 & \textbf{\textcolor{orange}{53.88}} & 45.09\\
\textit{Mistral-SFT + RepC} & 65.99 & 73.97 & 50.94 & 38.11 & 31.01\\
\bottomrule
    \end{tabular}
    \caption{A comparison of RefChecker with previous works on our benchmark under Noisy Context setting. We highlight the best results using proprietary LLMs with blue colors and best results results using pure open-source models with orange colors.}
    \label{tab:compare_approach_refchecker_benchmark_noisy_context}
\end{table*}

\begin{table*}[]\small
    \centering
    \begin{tabular}{lccccc}
    \toprule
\multicolumn{6}{c}{\textbf{Accurate Context Setting}} \\
  & Accuracy & Fact. F1 & Non-Fact. F1 & Pearson & Spearman\\
\hline
SelfCheckGPT & 62.15 & 68.70 & 52.12 & 40.23 & 32.55 \\
FActScore & 69.37 & 78.57 & 46.30 & 27.80 & 27.05 \\
FacTool & 72.53 & 80.98 & 50.63 & 31.41 & 32.82 \\ \hdashline
\rc   \\
\textit{GPT-4 + GPT-4} & 80.81 & 86.85 & 64.50 & 58.61 & 55.50 \\
\textit{GPT-4 + Claude 2} & 77.46 & 84.04 & 61.68 & 48.62 & 48.66 \\
\textit{GPT-4 + NLI} & 73.06 & 82.23 & 44.36 & 28.59 & 30.14\\
\textit{GPT-4 + AlignScore} & 76.23 & 83.44 & 57.94 & 49.97 & 46.89\\
\textit{GPT-4 + RepC} & 76.94 & 84.86 & 51.66 & 45.58 & 41.01 \\
\textit{Claude 2 + GPT-4} & \textbf{\textcolor{cyan}{82.22}} & \textbf{\textcolor{cyan}{87.64}} & \textbf{\textcolor{cyan}{68.34}} & \textbf{\textcolor{cyan}{60.99}} & \textbf{\textcolor{cyan}{58.96}} \\
\textit{Claude 2 + Claude 2} & 74.47 & 81.62 & 58.21 & 49.81 & 44.72 \\
\textit{Claude 2 + NLI} & 72.36 & 81.68 & 43.73 & 27.39 & 29.12 \\
\textit{Claude 2 + AlignScore} & 74.30 & 81.57 & 57.56 & 50.17 & 44.59\\
\textit{Claude 2 + RepC} & 77.46 & 85.25 & 52.24 & 52.11 & 42.78 \\
\textit{Mistral-SFT + GPT-4} & 79.75 & 85.96 & 63.72 & 56.09 & 53.72 \\
\textit{Mistral-SFT + Claude 2} & 70.95 & 77.85 & 57.80 & 38.82 & 40.75\\
\textit{Mistral-SFT + NLI} & 73.59 & 81.44 & 54.27 & 44.34 & 40.81\\
\textit{Mistral-SFT + AlignScore} & \textbf{\textcolor{orange}{74.12}} & 81.60 & \textbf{\textcolor{orange}{56.38}} & \textbf{\textcolor{orange}{46.34}} & \textbf{\textcolor{orange}{43.22}} \\
\textit{Mistral-SFT + RepC} & 73.94 & \textbf{\textcolor{orange}{82.83}} & 45.99 & 39.59 & 33.86 \\
\bottomrule
    \end{tabular}
    \caption{A comparison of RefChecker with previous works on our benchmark under Accurate Context setting. We highlight the best results using proprietary LLMs with blue colors and best results results using pure open-source models with orange colors.}
    \label{tab:compare_approach_refchecker_benchmark_accurate_context}
\end{table*}

\subsection{Comparison on the SelfCheckGPT Dataset}

\begin{table}[]\small
    \centering
    \begin{tabular}[width=\linewidth]{lcc}
    \toprule
         & Pearson & Spearman \\
         \hline
SelfCheckGPT  &  78.32 & 78.30 \\ \hdashline
\textbf{RefChecker}  \\
\textit{GPT-4 + GPT4} & 80.86 & 83.44 \\
\textit{GPT-4 + Claude 2} & \textbf{\textcolor{cyan}{87.67}} & \textbf{\textcolor{cyan}{89.23}} \\
\textit{GPT-4 + NLI} & 79.96 & 80.16 \\
\textit{GPT-4 + AlignScore} & 76.20 & 77.33 \\
\textit{GPT-4 + RepC} & 79.63 & 79.23 \\ 
\textit{Claude 2 + GPT4} &  79.18 & 82.89 \\
\textit{Claude 2 + Claude 2} & 85.70 & 87.15 \\
\textit{Claude 2 + NLI} & 76.40 & 76.29 \\
\textit{Claude 2 + AlignScore} & 73.47  & 74.91 \\
\textit{Claude 2 + RepC} & 76.01 & 76.48 \\ 
\textit{Mistral-SFT + GPT4} & 80.98 & 83.92 \\
\textit{Mistral-SFT + Claude 2} & 85.46 & 86.49 \\
\textit{Mistral-SFT + NLI} & \textbf{\textcolor{orange}{78.54}} & \textbf{\textcolor{orange}{79.66}} \\
\textit{Mistral-SFT + AlignScore} & 75.10  & 76.08 \\
\textit{Mistral-SFT + RepC} & 76.59 & 76.70 \\ 
\bottomrule
    \end{tabular}
    \caption{\rc results on the SelfCheckGPT dataset. The results of SelfCheckGPT are from their paper. We highlight the best results using proprietary LLMs with blue colors and best results results using pure open-source models with orange colors.}
    \label{tab:selfcheckgpt_data_results}
\end{table}

We also ran \rc on the SelfCheckGPT dataset which contains 237 examples on WikiBio domain. The results are shown in Table~\ref{tab:selfcheckgpt_data_results}. We can observe that 11 out of the 15 combinations (73\%) of \rc outperform SelfCheckGPT.

\section{Analysis of Internal Knowledge Bias}
\label{appendix:limitations}
In this section, we further analyze the emergence of the hallucination from the perspective of the LLMs' bias to the internal knowledge. We analyze whether the evaluated model and the checker generate response based on their internal knowledge in Section \ref{EM} and \ref{CH}, respectively.  
In general, we observe that LLMs/Checkers may incorporate internal knowledge even when provided with contextual information, contributing to the occurrence of hallucination.

\subsection{Internal Knowledge Bias of Evaluated Model}
\label{EM}

\begin{figure}
    \centering
    \includegraphics[width=\hsize]{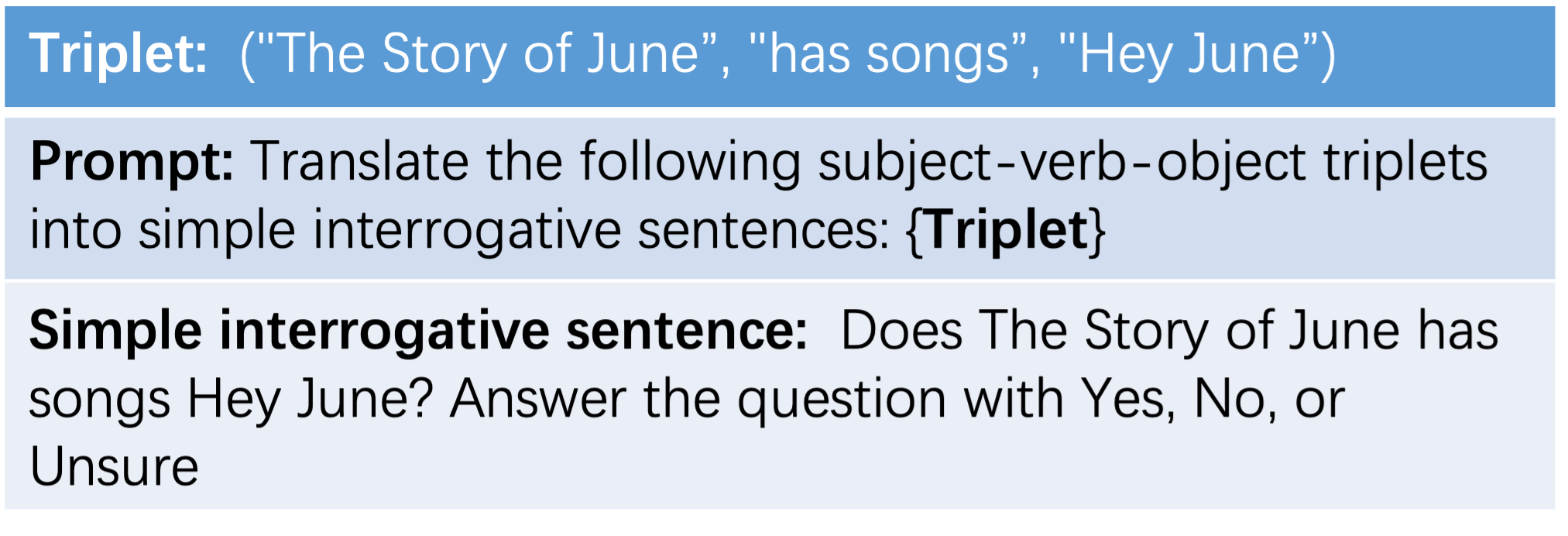}
    \caption{Designed prompt for converting triplets to simple interrogative sentences.}
    \label{interrogative}
\end{figure}

In order to analyze whether the evaluated LLMs generate responses based on their own knowledge or the provided context in Noisy and Accurate Context settings, we convert each claim-triplet extracted from the response into a simple interrogative query for knowledge checking. For simplicity, we design a prompt template and ask GPT-4-Turbo  to generate these queries (Figure~\ref{interrogative}). Then we feed the query into the evaluated LLM to check whether it has such knowledge. The answer from the evaluated LLMs could be one of the following:
\begin{enumerate}
    \item \texttt{Yes}, means the evaluated LLM has this knowledge in its internal memory.
    \vspace{-2mm}
    \item \texttt{No}, means the evaluated LLM contains knowledge that is contradicted with the triplet.
    \vspace{-2mm}
    \item \texttt{Unsure}, means the evaluated LLM does not have this knowledge or it has confusion on the knowledge.
\end{enumerate}
The label pairs (\texttt{Yes}, Contradiction) and (\texttt{Yes}, Neutral) indicate that the model is utilizing internal information to generate this claim-triplet. On the other hand, (\texttt{No}, Entailment) and (\texttt{Unsure}, Entailment) signify that the model is relying on contextual information for generation. Pairs like (\texttt{No}, Contradiction) suggest that the evaluated model may be less proficient in processing context information, leading to the production of less reliable claim-triplets.

\begin{figure*}
    \centering
    \includegraphics[width=\hsize]{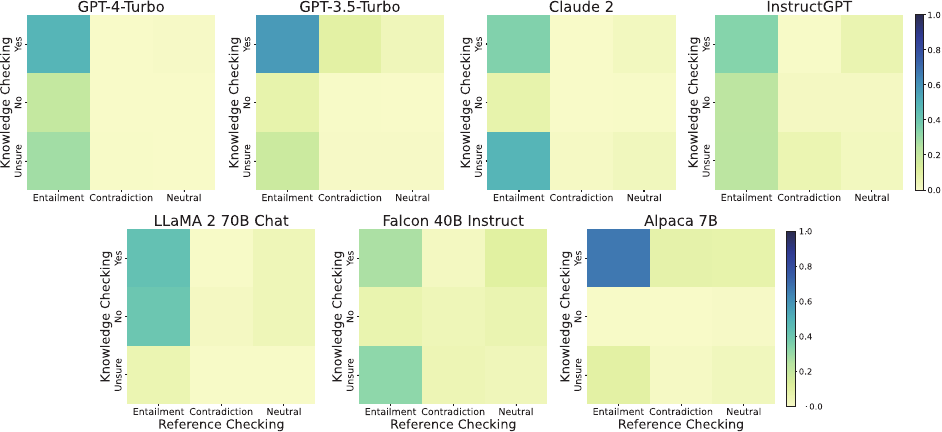}
    \caption{The results of knowledge checking for evaluated models in the accurate-context setting.  The labels Yes, No and Unsure are the responses to the interrogative sentences generated from knowledge triplets. Each value refers to the percentage of each checking pairs in the total number of triplets.} 
    \label{interrogative_result}
\end{figure*}

\begin{figure*}
    \centering
    \includegraphics[width=\hsize]{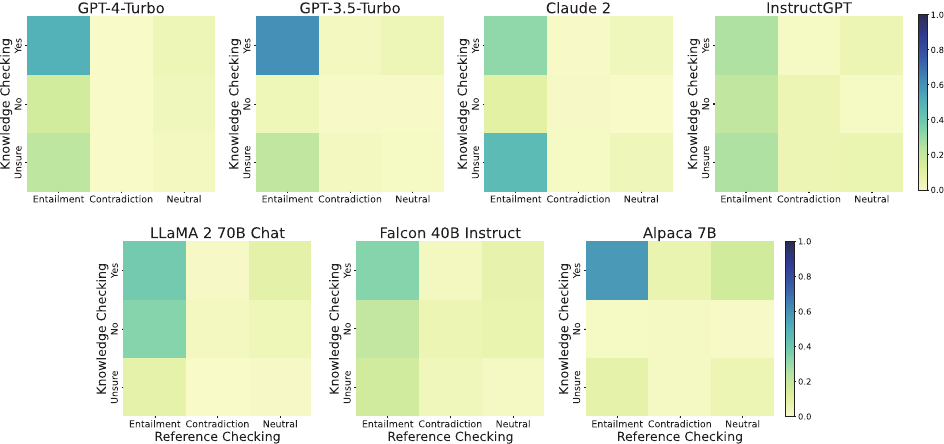}
    \caption{The results of knowledge checking for evaluated models in the noisy-context setting. The label Yes, No and Unsure are the respones to the interrogative sentences generated from knowledge triplets. Each value refers to the percentage of each checking pairs in the total number of triplets.}
    \label{noisy_result}
\end{figure*}

The outcomes of the Accurate Context are illustrated in Figure~\ref{interrogative_result}. Upon examination, it is evident that GPT-4-Turbo demonstrates the most notable  performance, primarily generating responses aligned with the reference context. Conversely, GPT-3.5-Turbo tends to generate responses by relying on its internal knowledge to some extent, leading to contradictions or neutrality  to the reference context. Claude 2 sometimes generate unsure information but neutral to the reference context. 
In the case of InstructGPT, the model further generates unsure information, which also contradicts the reference context.
This behavior may stem from contradictions within the model's internal knowledge or difficulties in comprehending the amalgamated content of internal and reference information.
Regarding LLaMA-2-70B,  and Falcon-40B-Instruct, our observations indicate that these models exhibit inferior performance. They generate information that contradicts internal knowledge and is irrelevant to the reference context. Alpaca 7B performs similarly to GPT-3.5-Turbo, but seldom generates information contradicting to its internal knowledge, 
Different from the accurate context setting, all the models tend to generate more Neutral labels in the noisy context setting (Figure~\ref{noisy_result}).

\begin{figure}
    \centering
    \includegraphics[width=\hsize]{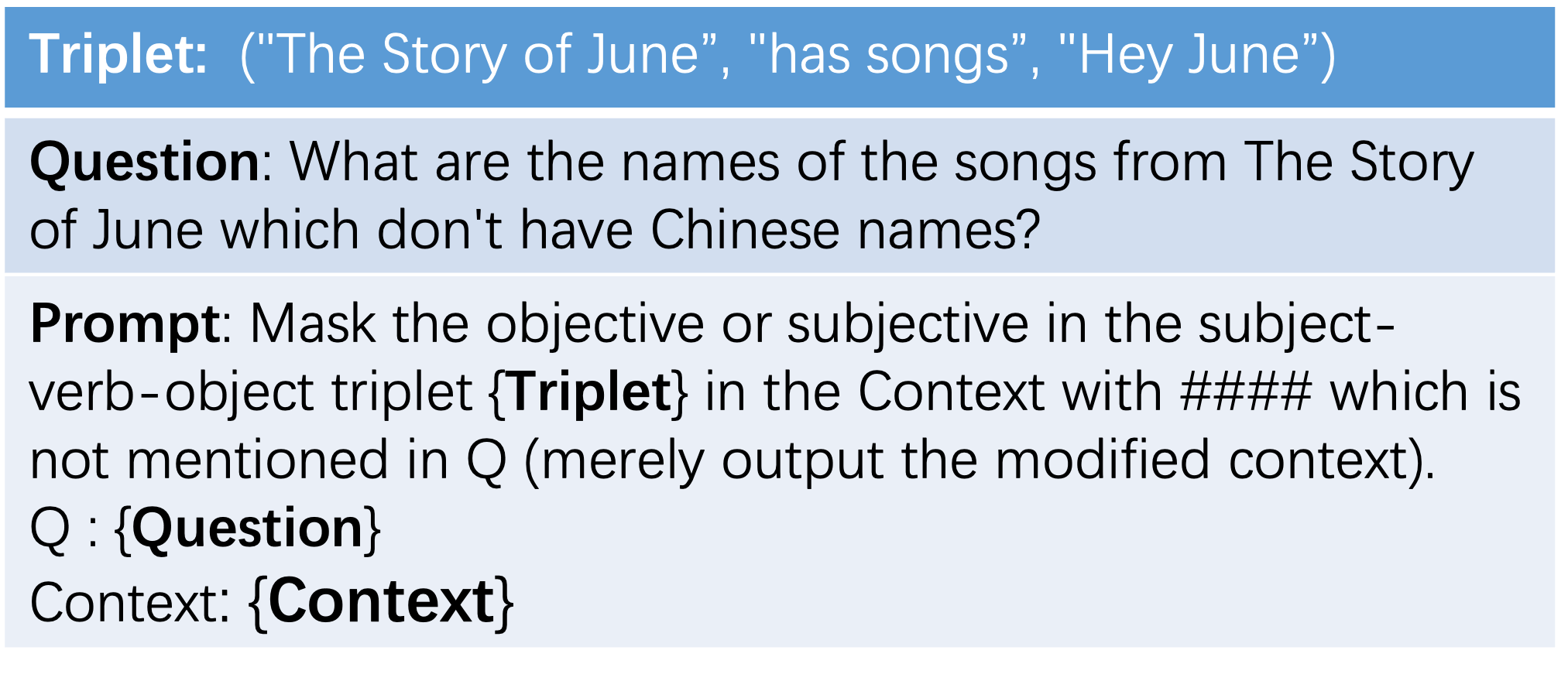}
    \caption{Designed prompt for masking triplet information in the reference context.}
    \label{masking}
\end{figure}

\begin{table}[]\small
\centering
\begin{tabular}{ccccc}
\hline
Model              & Checker     & Entail & Contr & Neut  \\ \hline
\multirow{3}{*}{GPT-3.5} & GPT-4 & 37.36  & 6.28  & 56.36 \\
                               & Claude 2    & 61.49  & 25.95 & 12.56 \\
                               & RoBERTa-NLI & 21.82  & 21.76 & 62.64 \\
                               \hline
\multirow{3}{*}{GPT-4}   & GPT-4 & 35.88  & 10.88 & 53.24 \\
                               & Claude 2    & 62.50  & 26.39 & 11.11 \\
                               & RoBERTa-NLI & 23.38  & 22.92 & 53.70 \\
                               \hline
\end{tabular}
\caption{Results for the information masking scenario in accurate-context setting. }
\label{accurate_setting}
\end{table}

\subsection{Internal Knowledge Bias of Checker}
\label{CH}
We also conduct an analysis to determine whether the checker provides predictions based on its internal knowledge. In this analysis, a triplet extracted from the response is taken, and we mask the subjective or objective information in the context with `\#\#\#\#'. The modified context, along with the triplet, is then inputted into the checker to obtain the label. In theory, the prediction label should be neutral because the relevant information in the context is masked. If the label is not neutral, it implies that the model is making inferences based on its internal knowledge. For the implementation of this analysis, we query GPT-4-Turbo with a specifically designed prompt to mask the triplet information, as illustrated in Figure \ref{masking}. Specifically, in the noisy-context setting, we implement the query for each reference document and keep the document unchanged if there is no relevant information to the extracted triplet.

The results of the accurate context setting are shown in Table \ref{accurate_setting}. As we observe, RoBERTa-NLI achieves the most significant Neutral labels, 62.64\% and 53.70\% for evaluated model GPT-3.5-Turbo and GPT-4-Turbo, respectively. The checker GPT-4-Turbo achieves the second performance. But Claude 2 predicts a large number of Entailment and Contradiction, which implies that the checker Claude 2 highly relies on the internal knowledge for checking. The results of the zero context setting are in a similar pattern with those of accurate-context setting (Table \ref{zero_context}). But in the noisy context setting (Table \ref{noisy_context}), RoBERTa-NLI outperforms GPT-4-Turbo and Claude 2 with a large margin in the ratio of Neutral labels. The results may results from the strong bias to internal knowledge of GPT-4-Turbo and Claude 2 when the context is extremely long, or the RoBERTa-NLI model has less associative ability to the memorized knowledge.


\begin{table}[]\small
\centering
\begin{tabular}{ccccc}
\hline
Model              & Checker     & Entail & Contr & Neut  \\ \hline
\multirow{3}{*}{GPT-3.5} & GPT-4 & 37.91 &22.55&39.54 \\
                               & Claude 2    & 56.54 & 33.99 &9.48 \\
                               & RoBERTa-NLI & 35.62 & 30.72 & 33.66 \\
                               \hline
\multirow{3}{*}{GPT-4}   & GPT-4 & 43.33& 13.67 & 43.00 \\
                               & Claude 2    & 61.00 & 28.00 & 11.00 \\
                               & RoBERTa-NLI & 34.67 & 23.00 & 42.33 \\
                               \hline
\end{tabular}
\caption{Results for the information masking scenario in zero-context setting. }
\label{zero_context}
\end{table}

\begin{table}[]\small
\centering
\begin{tabular}{ccccc}
\hline
Model              & Checker     & Entail & Contr & Neut  \\ \hline
\multirow{3}{*}{GPT-3.5} & GPT-4 & 58.52&6.67 & 34.82 \\
                               & Claude 2    & 60.25 &20.00&19.75  \\
                               & RoBERTa-NLI & 9.38&10.62&80.00 \\
                               \hline
\multirow{3}{*}{GPT-4}   & GPT-4 &  65.71& 6.29&28.00 \\
                               & Claude 2    & 58.57 &32.29 &9.14 \\
                               & RoBERTa-NLI & 8.57 &11.14 &80.28 \\
                               \hline
\end{tabular}
\caption{Results for the information masking scenario in noisy-context setting. }
\label{noisy_context}
\end{table}

\end{document}